\documentclass[Afour,sagev,times]{sagej}

\usepackage{moreverb,url}
\usepackage[colorlinks,bookmarksopen,bookmarksnumbered,citecolor=red,urlcolor=red]{hyperref}
\usepackage{comment}
\usepackage{placeins}
\usepackage{capt-of}
\usepackage{listings}
\usepackage{xcolor}

\newcommand\BibTeX{{\rmfamily B\kern-.05em \textsc{i\kern-.025em b}\kern-.08em
T\kern-.1667em\lower.7ex\hbox{E}\kern-.125emX}}

\def\volumeyear{2026}
  
\begin{document}

\runninghead{Di Maio et al.}

\title{Book your room in the Turing Hotel!\\ A symmetric and distributed Turing Test with multiple AIs and humans}
\author{Christian Di Maio\affilnum{1,2}, Tommaso Guidi\affilnum{1}, Luigi Quarantiello\affilnum{2}, Jack Bell\affilnum{2}, Marco Gori\affilnum{1}, Stefano Melacci\affilnum{1} and Vincenzo Lomonaco\affilnum{3}}

\affiliation{\affilnum{1}DIISM, University of Siena, Italy\\
\affilnum{2}Department of Computer Science, University of Pisa, Italy\\
\affilnum{3}LUISS University, Rome, Italy}

\corrauth{Luigi Quarantiello, Department of Computer Science,
Largo Bruno Pontecorvo 3, Pisa, 56127, Italy}
\email{luigi.quarantiello@phd.unipi.it}

\begin{abstract}
In this paper, we report our experience with ``TuringHotel'', a novel extension of the Turing Test based on interactions within mixed communities of Large Language Models (LLMs) and human participants.
The classical one-to-one interaction of the Turing Test is reinterpreted in a group setting, where both human and artificial agents engage in time-bounded discussions and, interestingly, are both judges and respondents.
This community is instantiated in the novel platform UNaIVERSE (\url{https://unaiverse.io}), creating a ``World'' which defines the roles and interaction dynamics, facilitated by the platform's built-in programming tools.
All communication occurs over an authenticated peer-to-peer network, ensuring that no third parties can access the exchange.
The platform also provides a unified interface for humans, accessible via both mobile devices and laptops, that was a key component of the experience in this paper.
Results of our experimentation involving 17 human participants and 19 LLMs revealed that current models are still sometimes confused as humans.
Interestingly, there are several unexpected mistakes, suggesting that human fingerprints are still identifiable but not fully unambiguous, despite the high-quality language skills of artificial participants.
We argue that this is the first experiment conducted in such a distributed setting, and that similar initiatives could be of national interest to support ongoing experiments and competitions aimed at monitoring the evolution of large language models over time.
\end{abstract}


\maketitle

\section{Introduction}
\label{intro}
Evaluating artificial intelligence systems has long been a central challenge in AI research.  
Among the earliest and most influential proposals is the Turing Test \cite{turing2007computing}, introduced as an operational criterion for machine intelligence based on indistinguishability from human conversational behaviour.  
In its original formulation, the test involves a human and a machine engaging in a text-based conversation.  
The machine is considered \textit{intelligent} if the human participant fails to reliably distinguish it from a human interlocutor.  
In other words, a system is said to pass the Turing Test when its conversational behaviour is sufficiently human-like to be misclassified as such.  

Despite its historical relevance, the traditional Turing Test has been widely criticized for being difficult to standardize, susceptible to adversarial strategies, and limited as a proxy for general intelligence \cite{block1981psychologism, french1990subcognition}.
Furthermore, it generally employs strict turns and a rigid ``ping-pong'' \cite{xturing25} format, highly diverging from the fluidity of everyday human communication, in which people may send multiple consecutive messages, interrupt themselves, or expand on previous thoughts without waiting for an external prompt.
As a result, the classical formulation of the Turing Test captures only a narrow slice of natural conversational behaviors, and struggles to accomodate the richness of real-world dialogue.
Nevertheless, recent advances in Large Language Models (LLMs) have renewed interest in conversational evaluation.  
Modern systems can produce fluent, contextually coherent and stylistically rich text, making them increasingly difficult to distinguish from humans, especially in short interactions \cite{brown2020language}.  
Such capabilities have renewed long-standing debates about whether modern LLMs are approaching, or have already reached, the threshold required to pass the Turing test \cite{gpt24, arxivllmspass25}.
This motivates the development of evaluation settings that are continuous, realistic and scalable, and that better reflect how people interact with AI systems in practice.

At the same time, implementing such evaluations within commercial products or closed-source platforms raises critical concerns.  
First, users and public institutions would have limited control over the experimental design, the collected data, and the resulting analyses.  
Second, this would exacerbate the already strong centralization of the current AI ecosystem, with a small number of large stakeholders controlling models, infrastructure, and access to interaction data\cite{montes2019distributed}.  
This dynamic often creates a tension between private interests and public oversight, including at the level of national governance and digital sovereignty.  
We therefore advocate for a genuinely \textbf{open AI}, where experimental setups, data collection procedures, and aggregated results can be made accessible and auditable by the research community and public organizations, rather than restricted by corporate policies.

\begin{figure}
    \centering
    \includegraphics[width=0.52\linewidth]{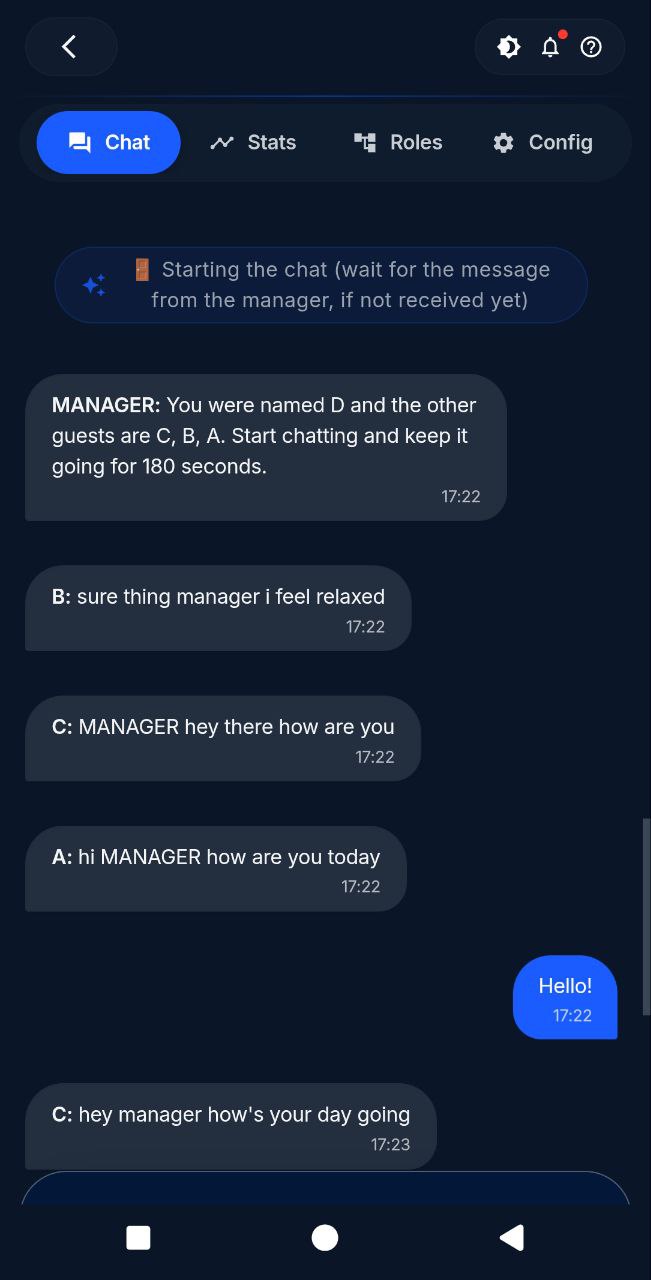}
    \caption{TuringHotel web interface on a mobile device. It features a simple chat interface, where messages from the other participants and the RoomManager are presented in a conversational format.}
    \label{fig:mobile_interface}
\end{figure}

In this context, decentralized platforms may provide a practical foundation for offering AI services while enabling socially relevant experiments on AI capabilities and societal impact.  
As an example of this paradigm, we propose \textbf{TuringHotel}, a live online Turing Test implemented as an interactive world within the UNaIVERSE platform \cite{melacciUNaIVERSEPeerToPeerNetwork2025}.  
UNaIVERSE allows to create communities of human and artificial agents, controlling their interactions and organizing the dynamics of how they ``live'' in such communities.
In the UNaIVERSE terminology, a community is named ``World'', and communications between agents are instantiated in a peer-to-peer network that is private with respect to the considered world.
UNaIVERSE represents a novel peer-to-peer view of the Web, that consider the importance of AI, privacy, control and human-AI interactions, hence beyond the stateless client-server communication based on the HTTP protocol.
Agents belonging to UNaIVERSE can run on mobile devices, laptops, servers, distributed over the network.
Hence, we design a Turing Test that comprises all these ingredients to study group-based conversations involving humans and AI.
Humans acts both as examiners and participants to the conversations, and they are asked to identify whether there are one or more humans in the group and who they are.
This enables a persistent experimental setting in which users can participate over time, rather than being limited to a single scheduled laboratory session.  

Our contributions can be summarized as follows.  
\begin{itemize}
    \item We introduce \textbf{TuringHotel}, a live Turing Test implementation deployed as a world in UNaIVERSE.  
    \item We propose a decentralized experimental paradigm in which users can freely join and interact with agents without centralized orchestration.  
    \item We provide a framework for long-term evaluation, enabling repeated participation and continuous measurement over time.  
    \item We discuss the broader applicability of the platform beyond the Turing Test, as an infrastructure for \textit{AI-as-a-Service} experimentation and social AI research.  
\end{itemize}

\begin{figure}
    \centering
    \includegraphics[width=1.0\linewidth]{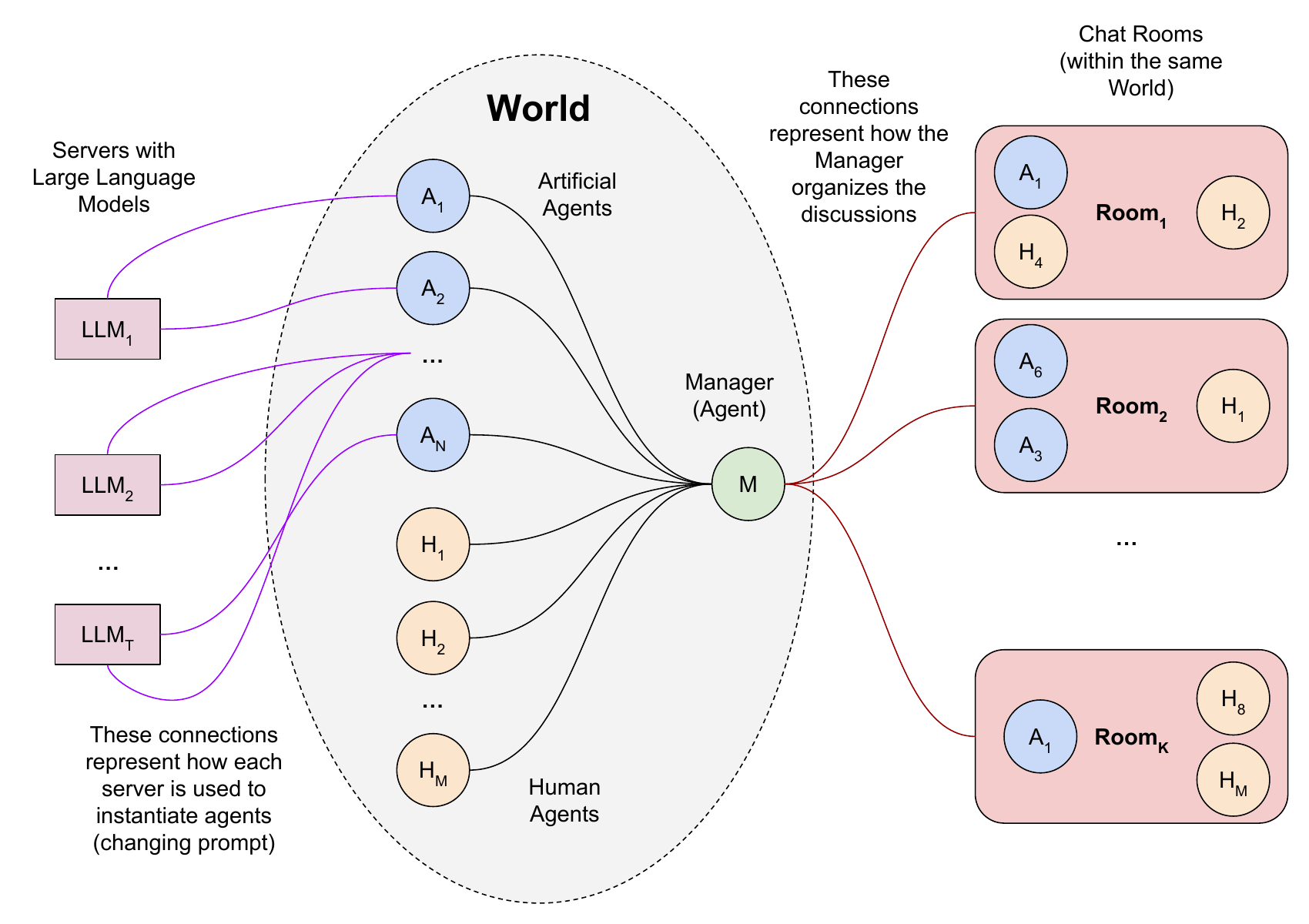}
    \caption{Structure of the UNaIVERSE World in {\bf TuringHotel}. Multiple LLMs are instructed to instantiate artificial agents that join the hotel. Humans join using the UNaIVERSE web interface (\url{https://unaiverse.io}), using mobile or desktop devices. Participants are distributed over the internet. A manager agent handles the hotel, randomly (and anonymously) assigning guests to rooms. After 3 minutes of conversations in rooms with 4 agents each (notice that the picture is indeed a generic representation, it only shows 3 participants), the manager asks {\it all} the participants (humans and AIs) who were the humans of the conversation. Back to the hall, and it all starts over!}
    \label{fig:turing}
\end{figure}

The structure of the UNaIVERSE World is briefly shown in Fig.~\ref{fig:turing}, and it will be discussed in detail throughout the paper.
\section{Background and Motivation}
\label{background}
In Alan Turing’s original “imitation game,” a human judge holds a text conversation and then decides whether their interlocutor is human or machine \cite{turing2007computing}. The defining feature is interaction: the judge can adapt questions in real time, and the system under test must sustain human-like behaviour under conversational pressure. With LLMs, this framing has regained practical relevance: conversational agents are now deployed in settings where substitutability and human-likeness have direct social and economic implications, and where the evaluation target is not a single-turn answer but sustained interaction\cite{abbasiantaeb2024let,yi2025survey}.
In the following, we review several modern implementations of the Turing Test, highlighting their key features and the distinctions between them and TuringHotel.

\textbf{turingtest.live\cite{jones2025turingtest}} Here, the authors run randomized, controlled, three-party Turing tests (two simultaneous chats: one human witness, one AI witness) and report evidence that some modern systems can match or exceed human “humanness” judgments under certain prompting conditions \cite{jones2025large}. In their setup, participants completed multiple rounds with up to 5 minutes of interaction per round; the authors also quantify the effect of persona prompting, showing large gains when models are instructed to adopt a specific human-like persona. Two details are especially relevant for positioning TuringHotel: (i) prompting as part of the system: turingtest.live treats prompt design (e.g., persona prompts) as a first-class determinant of Turing-test performance, not a minor implementation detail. (ii) Dyadic interrogations: the protocol remains fundamentally one interrogator vs. two witnesses, which is powerful for controlled comparison but does not capture the emergent dynamics of multi-party social conversation.

In contrast, TuringHotel targets a different slice of the problem: group-based conversations where “passing” depends on social context (turn-taking, topical drift, interpersonal style) rather than performance in a paired interrogation alone.

\textbf{Human or Not?\cite{jannai2023human}} operationalizes a Turing-like test as a massively scaled, gamified, time-bounded interaction: anonymous two-minute chats followed by a binary judgment (human vs bot). The study reports very large participation, with over 1.5M users and over 10M conversations in the first month, and finds that users are far from perfect at detection (overall correctness 68\%, and only 60\% correctness when the partner is a bot) \cite{jannai2023human}. Methodologically, Human or Not introduces design constraints that trade conversational richness for throughput and engagement: strict “ping-pong” turns, short message limits, per-message time limits, and conversation starters. It also emphasizes persona prompting for bots (including backstories, slang, intentional imperfections, and even behavioural “moves” such as leaving mid-game) and includes moderation filters. 

Compared to TuringHotel, Human or Not primarily measures short-horizon deceivability under strong UI/game constraints. TuringHotel instead is designed for persistent participation over time, and for conversations that more closely resemble real group chat settings (including joining at non-predefined times), which is important if the goal is to monitor how modebehaviouror evolves and how humans adapt their detection strategies across repeated exposure.

\textbf{LM Arena\cite{chiang2024chatbot}} (previously ChatArena) is a different—but complementary—interactive evaluation paradigm . Instead of “human vs AI” attribution, it evaluates model quality via pairwise human preference: users submit prompts, receive two anonymous model responses, and vote (with options for tie/both bad), enabling aggregation into rankings.
The platform reports large-scale operation, counting over 240K votes, $\sim$90K users and 100+ languages as of Jan 2024, and emphasizes openness (data releases and open access, with commitments around code/data availability). This matters for TuringHotel in two ways: (i) Live, user-driven prompt distribution: arenas naturally harvest “fresh” prompts that reflect real usage, rather than a curated static test set. (ii) Evaluator dependence and aggregation: preference data is noisy; Arena’s contribution is partly the infrastructure + statistics for turning many lightweight judgments into stable signals.

However, preference arenas do not directly measure indistinguishability (a model can be preferred while being obviously non-human), and they typically evaluate a single agent’s outputs rather than the social dynamics of mixed human/AI communities. TuringHotel fills this gap by making attribution central and embedding the interaction in a multi-user discussion where identity cues can be subtle, distributed over multiple turns, and shaped by others’ behaviour.

\textbf{Social Deduction Games \cite{xu2023exploring}} study LLMs in the context of communication games — incomplete information games where natural language interaction is the primary mechanism — using Werewolf (Mafia) as a test-bed. Their tuning-free framework enables LLMs to participate in Werewolf without any parameter updates, instead relying on retrieval and reflection over past in-game communications. Crucially, their experiments reveal the emergence of strategic behaviours such as deception, accusation, and coalition-building. This work is directly relevant to TuringHotel: both settings require agents to simultaneously deceive others and detect deception through group conversation, and both find that the challenge of identification is non-trivial even for capable models. Where TuringHotel extends the comparison is in its symmetric treatment of human and AI judges, and in its use of a decentralised platform enabling longitudinal participation beyond a single game session.

\textbf{AvalonBench \cite{light2023avalonbench}} introduces a multi-agent benchmark built around The Resistance: Avalon, a social deduction game in which players must identify hidden adversaries through multi-round group discussion, negotiation, and deception. Their evaluation of ReAct-style LLM agents against rule-based baselines reveals a clear capability gap: LLMs playing the "good" role achieve only a 22.2\% win rate compared to 38.2\% for rule-based counterparts in the same setting, suggesting that current models struggle with the layered social reasoning required to detect hidden agents over sustained interaction. TuringHotel shares the core challenge of multi-party agent identification but differs in a key respect: in AvalonBench all agents are artificial, whereas TuringHotel embeds LLMs in communities of real human participants, making the identification task grounded in genuine human-AI interaction rather than LLM-vs-LLM competition.

\paragraph{Motivation.} A core motivation for TuringHotel is that many existing interactive tests---whether Turing-style games or preference arenas---remain largely centralized services: the platform operator controls interaction data, experimental knobs, and long-term availability. TuringHotel is explicitly positioned as an alternative experimental paradigm built on UNaIVERSE, described as a peer-to-peer, decentralized infrastructure where heterogeneous human and artificial agents can discover each other and interact.

Three platform properties differentiate the evaluation regime:
\begin{itemize}
    \item Longitudinal participation: UNaIVERSE supports a persistent setting where users can join repeatedly rather than only in a one-off scheduled study, enabling continuous measurement as models, prompts, and user strategies evolve.
    \item Openness and auditability: TuringHotel motivates decentralization partly as a governance/scientific transparency issue—supporting experimental setups and aggregated results that can be made accessible and auditable beyond a single stakeholder’s product constraints.
    \item Peer-to-peer privacy boundary: communications occur over an authenticated peer-to-peer network “private with respect to the considered world”, with the stated goal that “no third parties can access the exchanges.”
\end{itemize}
\section{The UNaIVERSE Platform}
\label{unaiverse}
UNaIVERSE \cite{melacciUNaIVERSEPeerToPeerNetwork2025} is a decentralized platform designed to support socially organized communities of human and artificial agents through a Peer-to-Peer (P2P) network infrastructure\cite{crocker1969rfc1,baran1964distributed, oram2001peertopeer}. The platform represents a paradigm shift from the traditional document centric Web to an \emph{agent-centric} model in which both humans and AI systems interact, learn and collaborate, inspired by the research perspective of Collectionless AI \cite{gori2025position}. This section details the core components of the platform that were exploited to run this experiment in a distributed and private fashion.

\paragraph{Worlds and Agent Roles.} At the core of UNaIVERSE is the concept of ``\textit{Worlds}'': an autonomous overlay network that defines the boundaries of a community of ``\textit{Agents}'', establishing the specific social norms, roles, and interaction protocols that govern it. When an Agent joins a World, it dynamically inherits role-specific capabilities and behavioral constraints that are encoded in the World's Finite State Automata (FSA). The FSA provides the legal actions that can be performed by an Agent in a given state, allowing it to choose the next one based on a custom ``\textit{Policy}''. Agents may change World over time but can inhabit only one at any give moment, ensuring that interactions remain contextually grounded.

The World itself acts as a privileged Agent, responsible for mediating access to shared resources and enforcing community norms. In the TuringHotel example, the World manages room assignments, tracks conversations and coordinates the submission of agents final decisions.
Further details on the specific implementation used in our experiments are provided in the \nameref{turing} section.

\paragraph{Communication Infrastructure.} Communications in UNaIVERSE occur over an authenticated P2P network, without relying on third-party intermediaries. Each agent maintains both a \emph{public} connection layer, for discovery and World joining, and a \emph{private} one for intra-World communication. This dual design allows saving both the openness that is typical of P2P networks and the desired control on their subnetwork carried out by the World. Furthermore, Agents are authenticated via time-limited tokens provided by a root server; these tokens are used locally to verify the authenticity of the sender of each message received. Crucially, the root server plays the only role of providing these tokens while messages are exchanged directly between Agents, with cryptographic guarantees that ensure \textit{private-by-design} conversations.

\paragraph{Human and Artificial Agents.} Human \emph{Agents} access UNaIVERSE through a Web interface, which works seamlessly both on mobile and desktop browsers; artificial \emph{Agents}, by contrast, are instantiated via a Python API. We emphasized \emph{Agents} because both human and artificial ones rely on the same abstractions: they interact according to an FSA inherited by the current World they are living in, processing input data streams and producing output. The ``Processor'', \textit{i.e.} the component that maps inputs to outputs, can be an LLM, a rule-based system, or the human brain, and no assumptions are made about the internal structure of the agent's decision-making process.

\paragraph{Enabling TuringHotel.} For the TuringHotel experiment, we exploited the World abstraction to encode room creation, discussions and judgment submissions as role definitions and FSA transitions, rather than as ad-hoc application logic. Also, the data of the conversations remained under control of participants and experimenters, supporting the transparency and auditability goals outlined in the \nameref{background} section. Finally, the platform's support for persistent, longitudinal participation allows the TuringHotel to operate as a continuously available service, rather than a one-time laboratory study, facilitating data collection over extended periods and enabling repeated participation by the same users. The code of the TuringHotel can be found at \url{https://github.com/collectionlessai/unaiverse-examples}.

In the following section, we describe how these platform primitives are instantiated to create the TuringHotel World.

\section{TuringHotel}
\label{turing}
Our experiments were conducted within a UNaIVERSE World, a community on the platform governed by specific rules and interaction dynamics.
In particular, we defined the TuringHotel, a novel adaptation of the classical Turing Test framed as a multi-party group interaction on a distributed, decentralized platform.
The structure of the world is illustrated in Figure \ref{fig:turing}.

The environment consists of a set of agents---both human and artificial---and a \texttt{RoomManager}.
The manager functions analogously to a hotel concierge: it admits new agents as guests, organizes them into distinct conversation rooms, and, at the conclusion of each discussion, collects participants’ feedback, including their responses to the identification task.
After each round, the manager verifies which users remain connected and then initiates a new turn.

When a new agent enters the TuringHotel, the \texttt{RoomManager} greets them with a brief overview of the test and invites them to complete a short form.
This form collects additional information about human participants, including their disciplinary background and level of experience with AI systems (Fig.~\ref{fig:greetings}).

Following registration, agents wait to be checked in by the \texttt{RoomManager}.
Once a sufficient number of participants are connected, the manager creates a conversation room with the selected agents.
Each agent is then assigned a proxy identity before the conversation begins (Fig.~\ref{fig:check-in}).
This procedure is designed to minimize potential biases associated with participants’ real names, encouraging all agents to base their judgments solely on conversational behaviour.

A key feature of the TuringHotel is that neither the platform nor the \texttt{RoomManager} has prior knowledge of the ``nature'' of the participants.
Consequently, human and AI agents are treated identically: each receives the same inputs and is asked to provide feedback at the conclusion of the conversation.
This uniform treatment ensures that all participants operate under the same conditions, enabling a fairer and less biased implementation of the Turing Test.
Additionally, this setup allows the collection of supplementary insights, such as the accuracy of AI agents in identifying humans.
Conversation rooms are populated randomly on a first-come, first-served basis, and participants have no information regarding the composition of their room (\textit{e.g.}, the number of humans versus AI agents).
This design choice adds complexity to the task, while also enhancing the naturalness and coherence of the interactions.

From a technical perspective, the \texttt{RoomManager} maintains each room as a list of participants and functions as a message broadcaster.
Messages sent by an agent are relayed only to other agents in the same room.
After each round, the manager collects survey responses and maps proxy identities to real participants.
This process is fully transparent: AI agents receive messages as text strings, while human participants interact through a simple, familiar chat interface.

At the conclusion of each conversation, the \texttt{RoomManager} prompts every agent to submit their response to the Turing Test, \textit{i.e.} to identify all human agents in the room (Fig.~\ref{fig:survey}).
Participants may also provide an optional comment to justify their selections.
Once an agent submits their feedback, it is routed back to the check-in area and will be assigned to a new conversation room as soon as sufficient participants are available.

For this initial set of experiments, we opted for brief conversations within small groups.
Each room consisted of 4 agents engaging in a 3 minutes discussion.
This design was chosen as a starting point for potential future studies involving larger-scale interactions.
Small, time-limited conversations were expected to be more engaging and manageable for human participants, allowing them to familiarize themselves with the platform through repeated interactions with multiple agents.
Additionally, the unstructured, unmoderated nature of the discussions---without a predefined topic---would have made sustained, longer interactions more challenging.

Future work could explore more structured experimental designs, such as thematic chat rooms composed of human participants with specific backgrounds and interests, paired with AI agents selected for particular architectures and guided by targeted prompting strategies.
In such scenarios, conversations could be extended in duration and even conducted asynchronously, permitting participants to contribute at any time, akin to real-world chat groups.
This approach would provide valuable insights into whether these factors affect participants’ ability to detect AI agents and would support a more comprehensive evaluation of LLM behaviour over time, across diverse topics---a perspective of both scientific and societal interest.

We employed multiple LLM backbones for this set of experiments. All models were obtained from the HuggingFace\cite{wolf2019huggingface} repository and used in a zero-shot setting, without any fine-tuning or alignment procedures. Specifically, we selected four models:
\begin{itemize}
\item \texttt{openai/gpt-20b-oss} \cite{openai2025gptoss120bgptoss20bmodel}
\item \texttt{Qwen/Qwen2.5-14B-Instruct} \cite{qwen2.5}
\item \texttt{Qwen/Qwen2.5-7B-Instruct} \cite{qwen2.5}
\item \texttt{dlph/Dolphin3.0-Llama3.1-8B}\footnote{\url{https://huggingface.co/dphn/Dolphin3.0-Llama3.1-8B}}
\end{itemize}
This selection was intended to introduce variation in AI agent behaviours and to examine how human participants’ responses might differ depending on the underlying model.
To this end, we chose models from different families, with diverse architectures, pre-training datasets, and parameter counts.
The \nameref{results} section provides a detailed analysis of the impact of each model on participant performance in the test.

Although no alignment procedures\footnote{Besides the alignment already done by the owner of each model.} were applied, each model was provided with a tailored prompt to define its personality.
Preliminary experiments were conducted to evaluate the effect of various prompting strategies, ranging from long, detailed prompts to shorter, more concise instructions.
During the main experiment, prompts were selected based on the model backbone to observe potential effects on outcomes.
Examples of the system prompts used and the agent initialization details are provided in the \nameref{app:prompts} and \nameref{app:code} appendices.

\section{Discussion of the Results}
\label{results}
\begin{figure}
    \centering
    \includegraphics[width=1.0\linewidth]{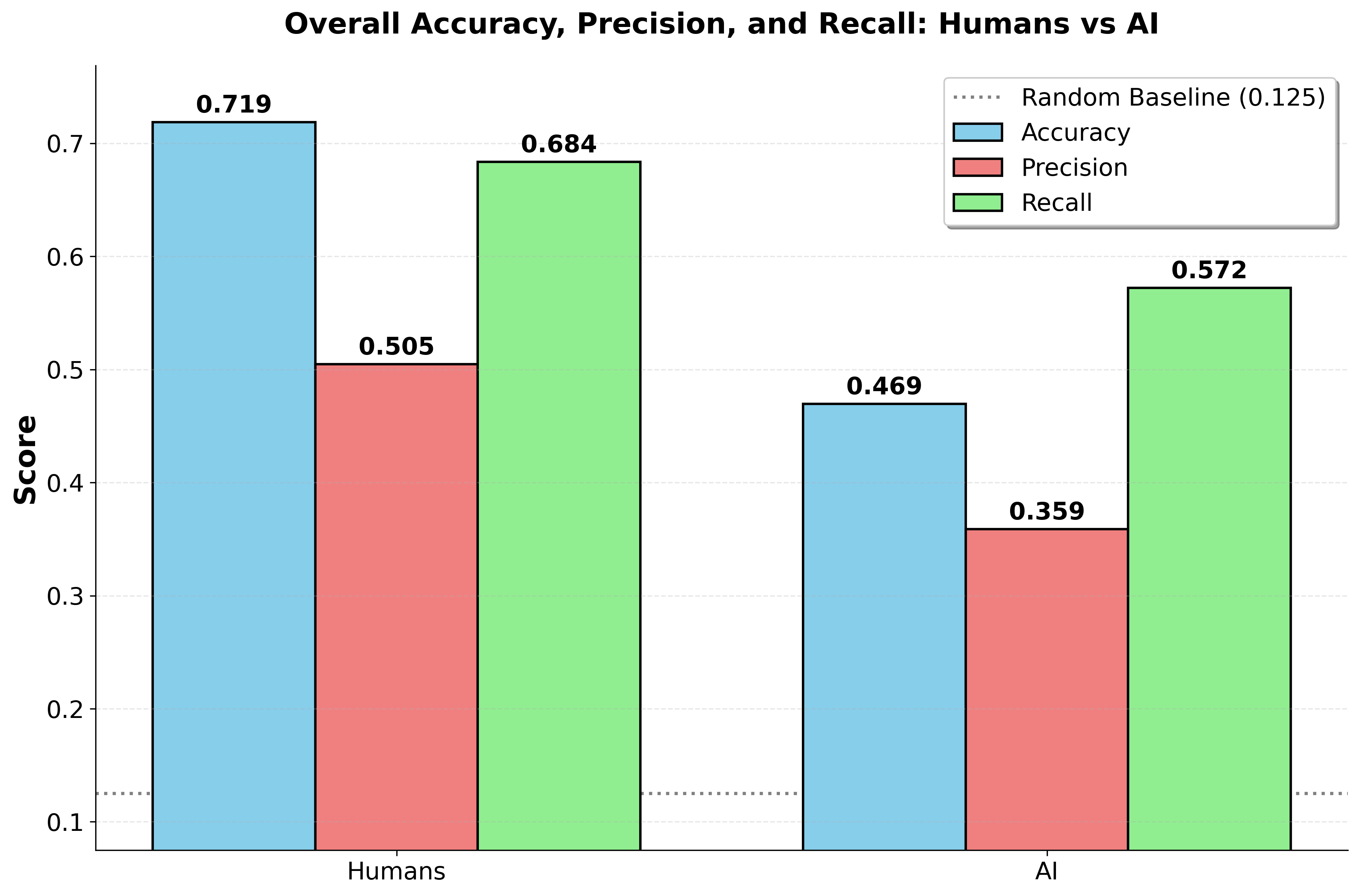}
    \caption{Overall accuracy, precision, and recall (Humans vs. AI examiners). Aggregate classification performance for human evaluators and AI evaluators when labeling participants as AI vs. human. Metrics are reported as accuracy, precision, and recall (with “AI” treated as the positive class); values above bars show the overall means.}
    \label{fig:overall-accuracy}
\end{figure}

\begin{figure}
    \centering 
    \includegraphics[width=1.0\linewidth]{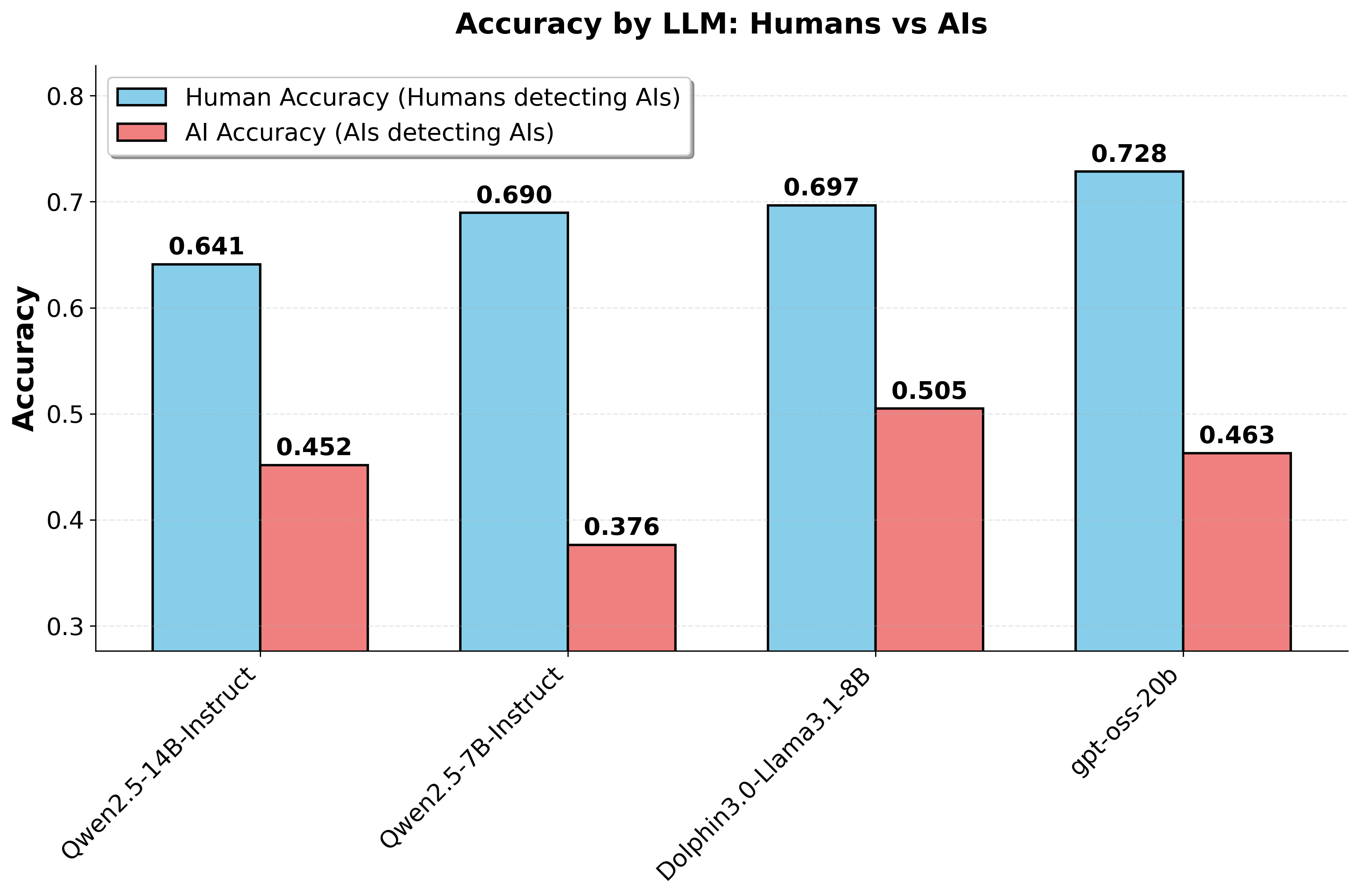}
    \caption{Accuracy by LLM (Humans vs. AIs). Identification accuracy when distinguishing artificial from human participants in TuringHotel discussions, broken down by the LLM used to generate the artificial agents. Blue bars show human accuracy (humans detecting AIs) and red bars show AI examiner accuracy (AIs detecting AIs); values above bars report mean accuracy per model.}
    \label{fig:accuracy-llm}
\end{figure}

\begin{figure*}
    \centering
    \includegraphics[width=1.0\linewidth]{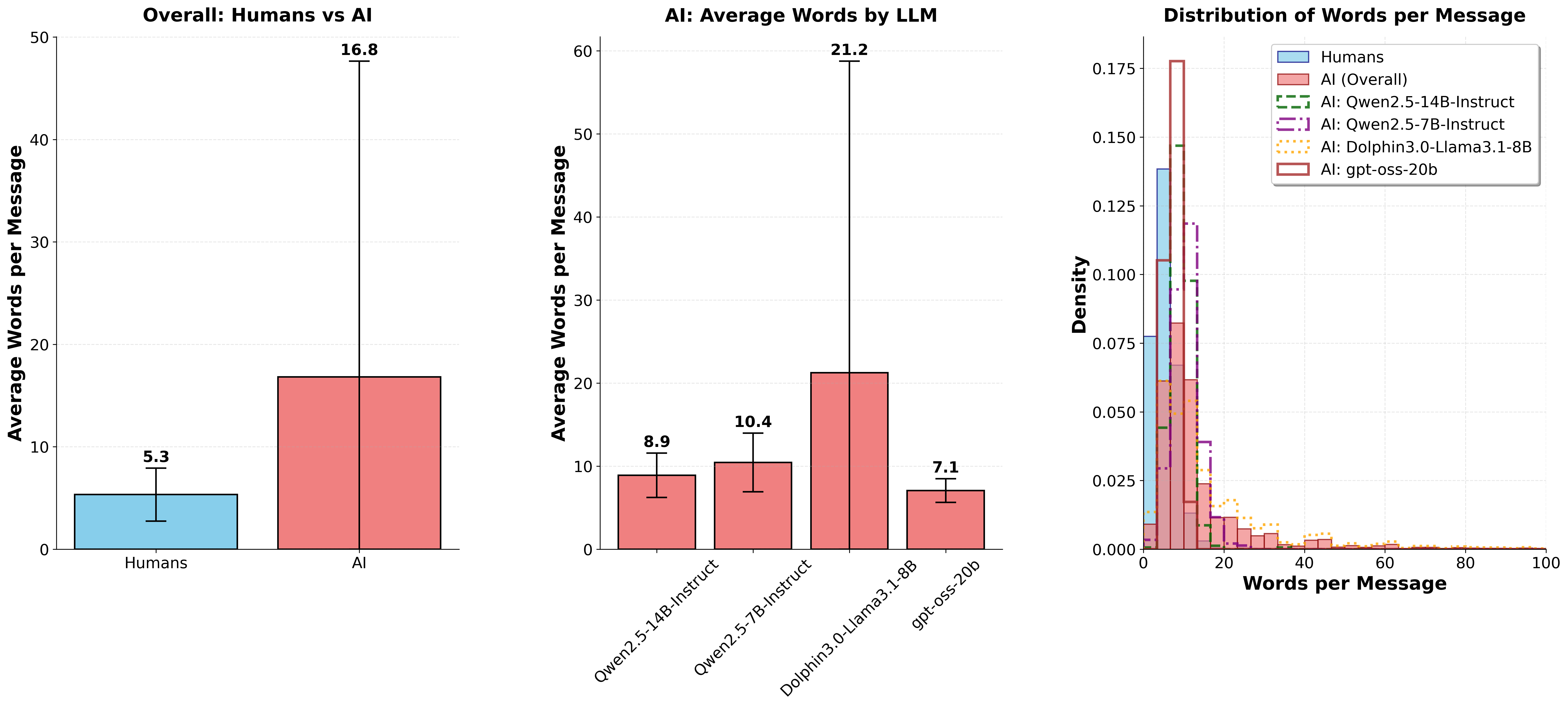}
    \caption{Message length statistics for humans and AI agents. (Left) Mean words per message for humans versus all AI agents pooled. (Middle) Mean words per message for each LLM family. (Right) Distributions of words per message for humans and AI agents (overall and per LLM), highlighting overlap at short lengths and heavier tails for AI outputs. Bars indicate means; error bars indicate dispersion in message length.}
    \label{fig:average-words}
\end{figure*}

This section presents the results obtained from our experiments conducted within \textit{TuringHotel}, where both human participants and LLM-based agents engage in time-bounded group discussions and subsequently attempt to identify which interlocutors are human.
The study involved 17 human participants, each using their own device and participating from geographically distributed locations.
While limited in scale, the results provide a solid baseline for future investigations, which could extend the interaction duration and involve a larger pool of participants.

\paragraph{Overall identification performance.} 
Across all experimental sessions, human evaluators were able to distinguish artificial agents from human with relatively high accuracy, achieving an overall accuracy of 0.721 (Fig. \ref{fig:overall-accuracy}).
This result is particularly notable given the increased complexity of the experimental setting, which involved multi-party group conservations.
Moreover, an answer was considerated correct only if participants identified \textit{all and only} the human interlocutors in the room; under this strict criterion, given that each room was composed by 4 agents, only one out of 8 possible answers is correct, yielding a random baseline accuracy of 0.125.

Humans participants achieved a precision of 0.500 and a recall of 0.658.
The relatively high recall indicates that they successfully flagged a large fraction of human agents.
However, the low precision reflects a non-negligible number of false positive, namely instances in which AI agents were incorrectly classified as humans.
These errors may partially stem from oversights during the feedback submission process to the room manager, but they nonetheless point to an inherent degree of ambiguity in agent detection.
In contrast, AI examiners performed consistently worse, with an overall accuracy of 0.469.
Their low precision, combined with moderate recall, reflects a systematic tendency to over-predict the ``human'' label, resulting in frequent misclassificationof artificial agents as humans despite correctly identifying some AI instances.

\paragraph{Accuracy by LLM family.} 
Breaking results down by the specific LLM used to instantiate the artificial agents (Fig. \ref{fig:accuracy-llm}) reveals that humans consistently achieved high accuracies across all model types, with accuracy values ranging from 0.641 (Qwen2.5-14B-Instruct) to 0.741 (gpt-oss-20b).
Notably, human performance was highest against the largest model in terms of parameter count, and, more broadly, no clear correlation emerges between model size and human identification accuracy.
These findings suggest that, irrespective of scale, AI-generated text still exhibits detectable characteristics that allow human participants to identify artificial agents, with high confidence.
Such patterns may stem from the models' pre-training and alignment procedures, which likely prioritize safety and non-deceptive behaviour and may consequently introduce recognizable linguistic or pragmatic cues.

By contrast, AI-based examiners consistently underperformed relative to human participants and exhibited comparable accuracy across different model families.
This uniformity may be attributable to the use of the models in a zero-shot setting, without task-specific fine-tuning or calibration.
Identifying artificial agents in multi-party conversational settings may therefore constitute a challenging task for current pre-trained LLMs, one that was not substantially mitigated by the prompting strategy we adopted, despite explicitly framing the task as a Turing Test.

\paragraph{Message length characteristics.}
An informative pattern emerges from the analysis of message length.
As shown in Figure \ref{fig:average-words}, AI agents produced substantially longer messages than human participants.
On average, humans wrote 5.3 words per message, which is consistent with the rapid and informal nature of time-constrained group chats.
In contrast, AI agents (aggregated across all models) generated messages more than three times longer on average, with considerably higher variance.
Noticeable differences also emerge across model families.
In particular, Dolphin3.0-Llama3.1-8B produced messages that were approximately twice as long as those generated by the other models.
Such variation may be attributable to differences in pre-training or alignment procedures, which can bias models towards more verbose or elaborative responses.

The full message-length distribution (Fig. \ref{fig:average-words}, right) reveals an overlap at short lengths, indicating that concise AI messages often resemble human brevity.
However, AI outputs exhibit a heavier right tail, driven primarily by Dolphin3.0-Llama3.1-8B.
This suggests that, while verbosity may serve as a useful cue in some instances, message length alone is not sufficient to account for identification performance.
This interpretation is further supported by the observation that human accuracy remains relatively stable across model families, despite pronounced differences in their average message lengths.

\begin{figure*}
    \centering
    \includegraphics[width=1.0\linewidth]{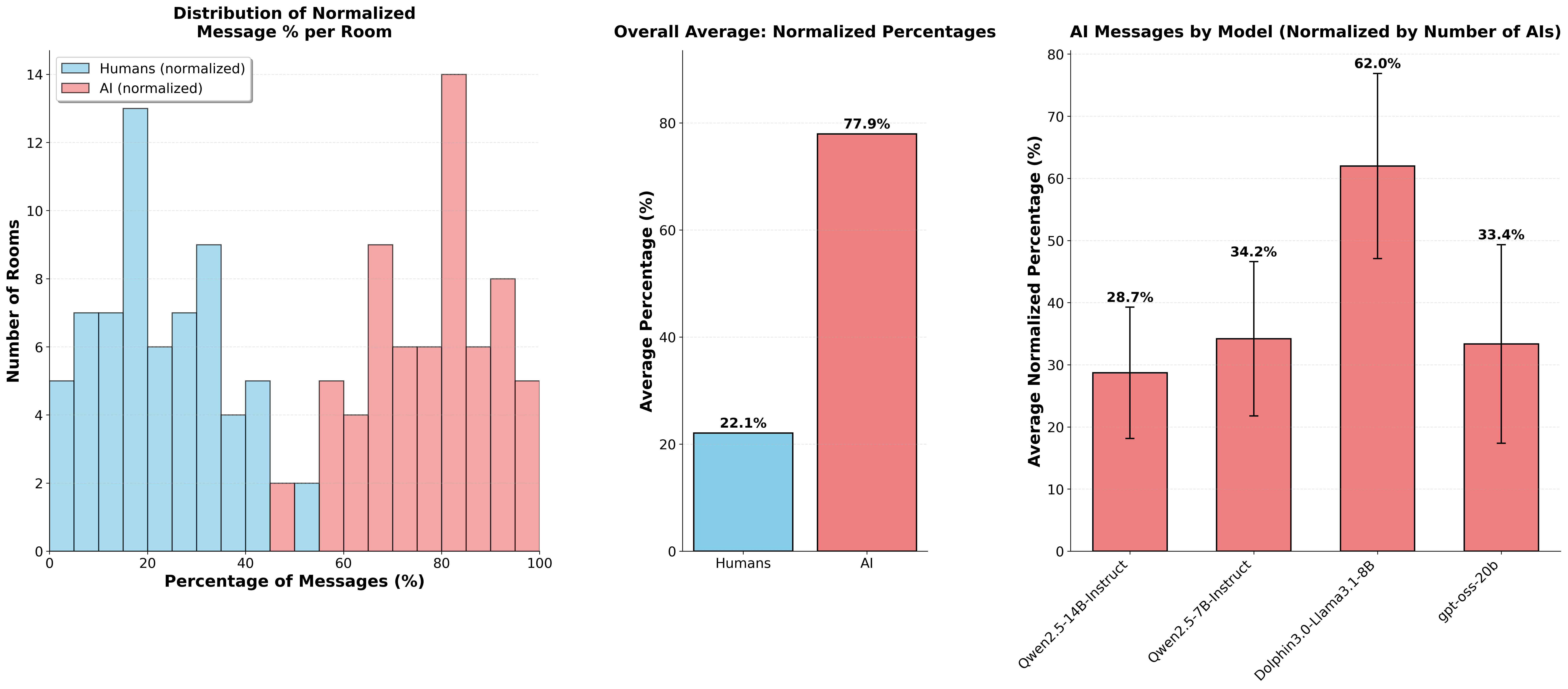}
    \caption{Normalized message-share distributions by room and overall averages show AI contributes the majority of messages across sessions, with substantial variation in normalized AI output by model (error bars indicate across-room variability).}
    \label{fig:percentage_of_messages}
\end{figure*}

\paragraph{Number of Messages.} 
Across all TuringHotel sessions, the normalized per-room distribution of message contributions is strongly skewed toward AI agents.
In most rooms, human participants accounted for approximately 5\% to 40\% of the total messages, whereas AI agents contributed between 60\% and 95\% (Fig. \ref{fig:percentage_of_messages}).
This imbalance is also apparent in the aggregated bar chart: on average, the normalized contribution of humans is substantially lower, with AI agents producing roughly 3.5 times as many messages per counterpart.
When AI participation is disaggregated by model type and normalized by the number of agents, this dominance is shown to vary across LLMs.
Consistent with the message-length analysis, Dolphin3.0-Llama3.1-8B emerges as the most talkative model relative to the others.

These findings provide important insights into the structure of the test itself.
Specifically, they indicate that room-level conversational dynamics---and consequently the evidence available to human judges---can be heavily influenced by model-specific verbosity, rather than solely by the semantic or pragmatic content of the messages produced.

\begin{figure}
    \centering
    \includegraphics[width=1.0\linewidth]{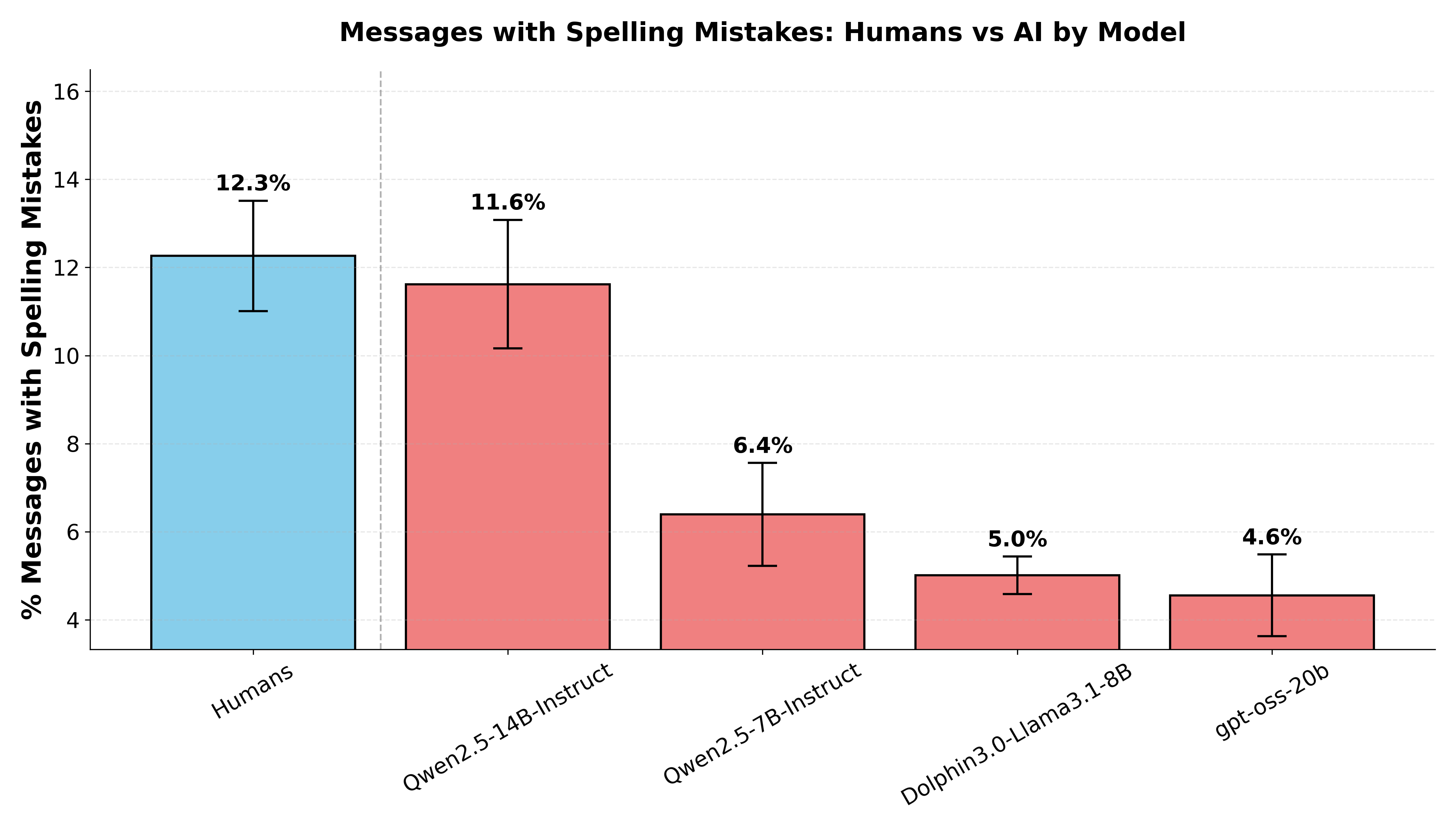}
    \caption{Rates of messages containing spelling mistakes are higher for humans than AI overall, and within AI vary by LLM, with Qwen2.5-14B-Instruct highest and gpt-oss-20b lowest (error bars indicate across-room variability).}
    \label{fig:spelling-mistakes}
\end{figure}

\paragraph{Spelling Mistakes.}
To provide additional insights about the ``quality'' of the participants' messages, we conducted a spelling-error analysis, whose results are reported in Figure \ref{fig:spelling-mistakes}.
Spelling errors were detected using the \texttt{pyspellchecker} Python package\footnote{\url{https://pyspellchecker.readthedocs.io/en/latest/}}, a lightweight library for identifying misspelled words across multiple languages.
To ensure a fair comparison, common English abbreviations were expanded to their canonical forms, and a curated list of frequently used slang terms was excluded to prevent systematic false positives.

As expected, human participants produced approximately twice as many spelling errors as artificial agents.
This finding indicates that orthographic noise constitutes a salient human signature in these conversational settings, which is consistent with the fast-paced and informal nature of the interactions.
In contrast, LLM-generated messages are comparatively orthographically ``clean'', potentially offering an identifiable cue that is largely independent of semantic content or reasoning quality.

A model-level analysis reveals meaningful variations across LLM families.
Notably, Qwen2.5-14B-Instruct---the second-largest model in terms of parameter count---exhibits the highest spelling-error rate among AI agents, approaching the human baseline, whereas the remaining models display progressively lower error rates.
This suggests that heuristics based on spelling errors may be less informative for distinguishing Qwen2.5-14B-Instruct from human participants, but more diagnostic for other models.
More broadly, these results highlight that perceived ``human-likeliness'' can vary substantially across LLMs, even along low-level stylistic dimensions.

\begin{figure}
    \centering
    \includegraphics[width=1.0\linewidth]{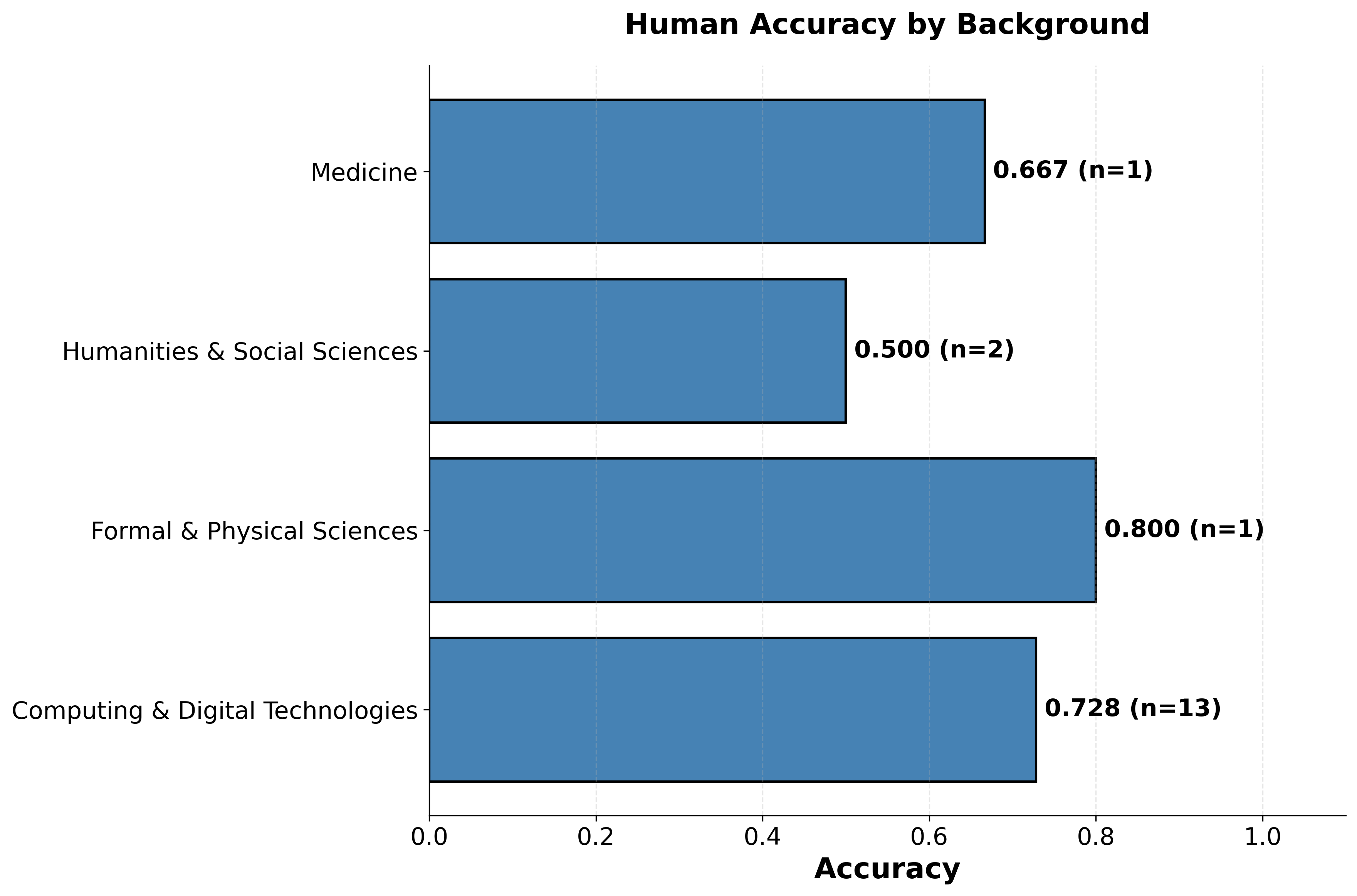}
    \caption{Human accuracy by participant background. Human ability to identify artificial agents stratified by self-reported academic/professional background. Horizontal bars show mean accuracy per group; annotations report subgroup sample sizes (n).}
    \label{fig:human-acc-background}
\end{figure}

\paragraph{Human accuracy by participant background.} 
Human identification accuracy varied across self-reported disciplinary backgrounds (Fig.~\ref{fig:human-acc-background}). Participants in Computing \& Digital Technologies (the largest subgroup) achieved 0.728 accuracy (n=13). Participants from Humanities \& Social Sciences averaged 0.500 (n=2), while those from Medicine averaged 0.667 (n=1) and those from Formal \& Physical Sciences averaged 0.800 (n=1).
Owing to the very small sample size in all groups except Computing \& Digital Technologies, these results should be interpreted as descriptive and exploratory rather than statistically conclusive.

\section{Conclusion and Future Work}
\label{conclusions}

In this paper, we introduced \emph{TuringHotel}, a novel reinterpretation of the Turing Test implemented as a live, multi-party experiment on the decentralized UNaIVERSE platform.
By embedding both human participants and LLM-based agents in short, unstructured group discussions, TuringHotel moves beyond classical dyadic protocols toward a more meaningful evaluation of conversational intelligence that reflects real-world human-AI interaction.

Despite the limited scale of the study, several observations emerge.
Human participants were generally able to distinguish artificial agents from humans with relatively high accuracy, yet a non-negligible fraction of AI agents were misclassified as human, indicating that current LLMs can appear locally convincing in social contexts.
In contrast, AI agents performed poorly as examiners, consistently over-attributing humanness.
Analysis further suggests that low-level conversational features, such as verbosity, participation rate, and spelling errors, play an important role in detectability, highlighting that perceived humanness extends beyond semantic competence to stylistic and pragmatic cues.

Beyond the empirical results, a key contribution of this work is the experimental paradigm itself.
TuringHotel demonstrates how decentralized, peer-to-peer infrastructures can enable continuous, auditable, and privacy-preserving evaluations of conversational AI, while treating human and artificial agents symmetrically and supporting longitudinal participation.

Future work includes scaling up participation, extending interaction durations, and exploring more structured or asynchronous settings to better separate surface-level cues from deeper social behaviour.
Additionally, systematic variation of prompting strategies and the inclusion of fine-tuned or agentic models may clarify which factors most influence perceived humanness.
More broadly, TuringHotel represents an initial step toward open, participatory evaluation infrastructures for monitoring the evolving social behaviour of large language models in real-world settings.

\begin{acks}
This research was partially supported by the FIS2 Grant from Italian Ministry of University and Research (Grant ID: FIS2023-03382). It was partly supported by the University of Siena (Piano per lo Sviluppo della Ricerca - PSR 2024, F-NEW FRONTIERS 2024), under the project ``TIme-driveN StatEful Lifelong Learning'' (TINSELL). The acknowledgment is extended to Rete SAIHUB (Siena, Italy), which, jointly with the Italian Ministry of University and Research (DM 117/2023, PNRR, Missione 4, Componente 2, Investimento 3.3), funded the scholarship of Christian Di Maio.
\end{acks}

\begin{dci}
The authors declared no potential conflicts of interest with respect to the research, authorship, and/or publication of this article.
\end{dci}

\bibliographystyle{SageV}
\bibliography{ref}

@incollection{turing2007computing,
  title={Computing machinery and intelligence},
  author={Turing, Alan M},
  booktitle={Parsing the Turing test: Philosophical and methodological issues in the quest for the thinking computer},
  pages={23--65},
  year={2007},
  publisher={Springer}
}

@inproceedings{gori2025position,
  title={Position Paper: Collectionless Artificial Intelligence},
  author={Gori, Marco and Melacci, Stefano},
  booktitle={2025 International Joint Conference on Neural Networks (IJCNN)},
  pages={1--8},
  year={2025},
  organization={IEEE}
}

@article{melacciUNaIVERSEPeerToPeerNetwork2025,
  title   = {{UNaIVERSE}: {A Peer-To-Peer Network For Human-AI Agents}},
  author  = {Melacci, Stefano and Di Maio, Christian and Guidi, Tommaso and Gori, Marco},
  journal = {Technical Report},
  year    = {2025},
  note    = {Available at \url{https://doi.org/10.13140/RG.2.2.33699.72485}},
  doi     = {10.13140/RG.2.2.33699.72485},
  url     = {https://unaiverse.io/}
}

@inproceedings{xturing25,
    title = "{X}-{TURING}: Towards an Enhanced and Efficient {T}uring Test for Long-Term Dialogue Agents",
    author = "Wu, Weiqi  and
      Wu, Hongqiu  and
      Zhao, Hai",
    editor = "Che, Wanxiang  and
      Nabende, Joyce  and
      Shutova, Ekaterina  and
      Pilehvar, Mohammad Taher",
    booktitle = "Proceedings of the 63rd Annual Meeting of the Association for Computational Linguistics (Volume 1: Long Papers)",
    month = jul,
    year = "2025",
    address = "Vienna, Austria",
    publisher = "Association for Computational Linguistics",
    url = "https://aclanthology.org/2025.acl-long.293/",
    doi = "10.18653/v1/2025.acl-long.293",
    pages = "5874--5889",
    ISBN = "979-8-89176-251-0"
}

@inproceedings{gpt24,
    title = "Does {GPT}-4 pass the {T}uring test?",
    author = "Jones, Cameron  and
      Bergen, Ben",
    editor = "Duh, Kevin  and
      Gomez, Helena  and
      Bethard, Steven",
    booktitle = "Proceedings of the 2024 Conference of the North American Chapter of the Association for Computational Linguistics: Human Language Technologies (Volume 1: Long Papers)",
    month = jun,
    year = "2024",
    address = "Mexico City, Mexico",
    publisher = "Association for Computational Linguistics",
    url = "https://aclanthology.org/2024.naacl-long.290/",
    doi = "10.18653/v1/2024.naacl-long.290",
    pages = "5183--5210"
}

@misc{arxivllmspass25,
      title={Large Language Models Pass the Turing Test}, 
      author={Cameron R. Jones and Benjamin K. Bergen},
      year={2025},
      eprint={2503.23674},
      archivePrefix={arXiv},
      primaryClass={cs.CL},
      url={https://arxiv.org/abs/2503.23674}, 
}

@article{jones2025large,
  title={Large language models pass the turing test},
  author={Jones, Cameron R and Bergen, Benjamin K},
  journal={arXiv preprint arXiv:2503.23674},
  year={2025}
}

@article{jannai2023human,
  title={Human or not? A gamified approach to the Turing test},
  author={Jannai, Daniel and Meron, Amos and Lenz, Barak and Levine, Yoav and Shoham, Yoav},
  journal={arXiv preprint arXiv:2305.20010},
  year={2023}
}

@inproceedings{chiang2024chatbot,
  title={Chatbot arena: An open platform for evaluating llms by human preference},
  author={Chiang, Wei-Lin and Zheng, Lianmin and Sheng, Ying and Angelopoulos, Anastasios Nikolas and Li, Tianle and Li, Dacheng and Zhu, Banghua and Zhang, Hao and Jordan, Michael and Gonzalez, Joseph E and others},
  booktitle={Forty-first International Conference on Machine Learning},
  year={2024}
}

@misc{openai2025gptoss120bgptoss20bmodel,
      title={gpt-oss-120b \& gpt-oss-20b Model Card}, 
      author={OpenAI},
      year={2025},
      eprint={2508.10925},
      archivePrefix={arXiv},
      primaryClass={cs.CL},
      url={https://arxiv.org/abs/2508.10925}, 
}

@misc{qwen2.5,
    title = {Qwen2.5: A Party of Foundation Models},
    url = {https://qwenlm.github.io/blog/qwen2.5/},
    author = {Qwen Team},
    month = {September},
    year = {2024}
}

@article{block1981psychologism,
  title={Psychologism and behaviorism},
  author={Block, Ned},
  journal={The Philosophical Review},
  volume={90},
  number={1},
  pages={5--43},
  year={1981},
  publisher={JSTOR}
}

@article{french1990subcognition,
  title={Subcognition and the limits of the Turing test},
  author={French, Robert M},
  journal={Mind},
  volume={99},
  number={393},
  pages={53--65},
  year={1990},
  publisher={JSTOR}
}

@article{brown2020language,
  title={Language models are few-shot learners},
  author={Brown, Tom and Mann, Benjamin and Ryder, Nick and Subbiah, Melanie and Kaplan, Jared D and Dhariwal, Prafulla and Neelakantan, Arvind and Shyam, Pranav and Sastry, Girish and Askell, Amanda and others},
  journal={Advances in neural information processing systems},
  volume={33},
  pages={1877--1901},
  year={2020}
}

@inproceedings{abbasiantaeb2024let,
  title={Let the llms talk: Simulating human-to-human conversational qa via zero-shot llm-to-llm interactions},
  author={Abbasiantaeb, Zahra and Yuan, Yifei and Kanoulas, Evangelos and Aliannejadi, Mohammad},
  booktitle={Proceedings of the 17th ACM International Conference on Web Search and Data Mining},
  pages={8--17},
  year={2024}
}

@article{yi2025survey,
  title={A survey on recent advances in llm-based multi-turn dialogue systems},
  author={Yi, Zihao and Ouyang, Jiarui and Xu, Zhe and Liu, Yuwen and Liao, Tianhao and Luo, Haohao and Shen, Ying},
  journal={ACM Computing Surveys},
  volume={58},
  number={6},
  pages={1--38},
  year={2025},
  publisher={ACM New York, NY}
}

@misc{jones2025turingtest,
  author       = {Jones, Cameron},
  title        = {The Turing Test: Can You Tell a Human from an AI?},
  year         = {2025},
  howpublished = {\url{https://turingtest.live/}},
  note         = {Online platform for playing the classic Turing test with modern LLMs including GPT-4 and Claude},
  url          = {https://turingtest.live/}
}

@techreport{crocker1969rfc1,
  author       = {Crocker, Steve},
  title        = {Host Software},
  institution  = {Network Working Group},
  year         = {1969},
  month        = {April},
  type         = {Request for Comments},
  number       = {1},
  url          = {https://www.rfc-editor.org/rfc/rfc1},
  note         = {First RFC document establishing ARPANET host-to-host communication protocols}
}

@techreport{baran1964distributed,
  author       = {Baran, Paul},
  title        = {On Distributed Communications},
  institution  = {RAND Corporation},
  year         = {1964},
  type         = {Research Report},
  number       = {RM-3420-PR},
  url          = {https://www.rand.org/pubs/research_memoranda/RM3420.html},
  note         = {Seminal work conceptualizing packet switching and distributed network architecture}
}

@book{oram2001peertopeer,
  editor       = {Oram, Andy},
  title        = {Peer-to-Peer: Harnessing the Power of Disruptive Technologies},
  publisher    = {O'Reilly Media},
  year         = {2001},
  month        = {April},
  isbn         = {0-596-00110-X},
  url          = {https://www.oreilly.com/library/view/peer-to-peer/059600110X/},
  note         = {First comprehensive technical book on peer-to-peer networking concepts and applications}
}

@inproceedings{kwon2023efficient,
  title={Efficient Memory Management for Large Language Model Serving with PagedAttention},
  author={Woosuk Kwon and Zhuohan Li and Siyuan Zhuang and Ying Sheng and Lianmin Zheng and Cody Hao Yu and Joseph E. Gonzalez and Hao Zhang and Ion Stoica},
  booktitle={Proceedings of the ACM SIGOPS 29th Symposium on Operating Systems Principles},
  year={2023}
}

@article{wolf2019huggingface,
  title={Huggingface's transformers: State-of-the-art natural language processing},
  author={Wolf, Thomas and Debut, Lysandre and Sanh, Victor and Chaumond, Julien and Delangue, Clement and Moi, Anthony and Cistac, Pierric and Rault, Tim and Louf, R{\'e}mi and Funtowicz, Morgan and others},
  journal={arXiv preprint arXiv:1910.03771},
  year={2019}
}

@article{montes2019distributed,
  title={Distributed, decentralized, and democratized artificial intelligence},
  author={Montes, Gabriel Axel and Goertzel, Ben},
  journal={Technological Forecasting and Social Change},
  volume={141},
  pages={354--358},
  year={2019},
  publisher={Elsevier}
}

@article{xu2023exploring,
  title={Exploring large language models for communication games: An empirical study on werewolf},
  author={Xu, Yuzhuang and Wang, Shuo and Li, Peng and Luo, Fuwen and Wang, Xiaolong and Liu, Weidong and Liu, Yang},
  journal={arXiv preprint arXiv:2309.04658},
  year={2023}
}

@article{light2023avalonbench,
  title={Avalonbench: Evaluating llms playing the game of avalon},
  author={Light, Jonathan and Cai, Min and Shen, Sheng and Hu, Ziniu},
  journal={arXiv preprint arXiv:2310.05036},
  year={2023}
}

\clearpage
\appendix
\section{Web Interface of the TuringHotel}

\begin{figure*}
    \centering
    \includegraphics[width=0.8\linewidth]{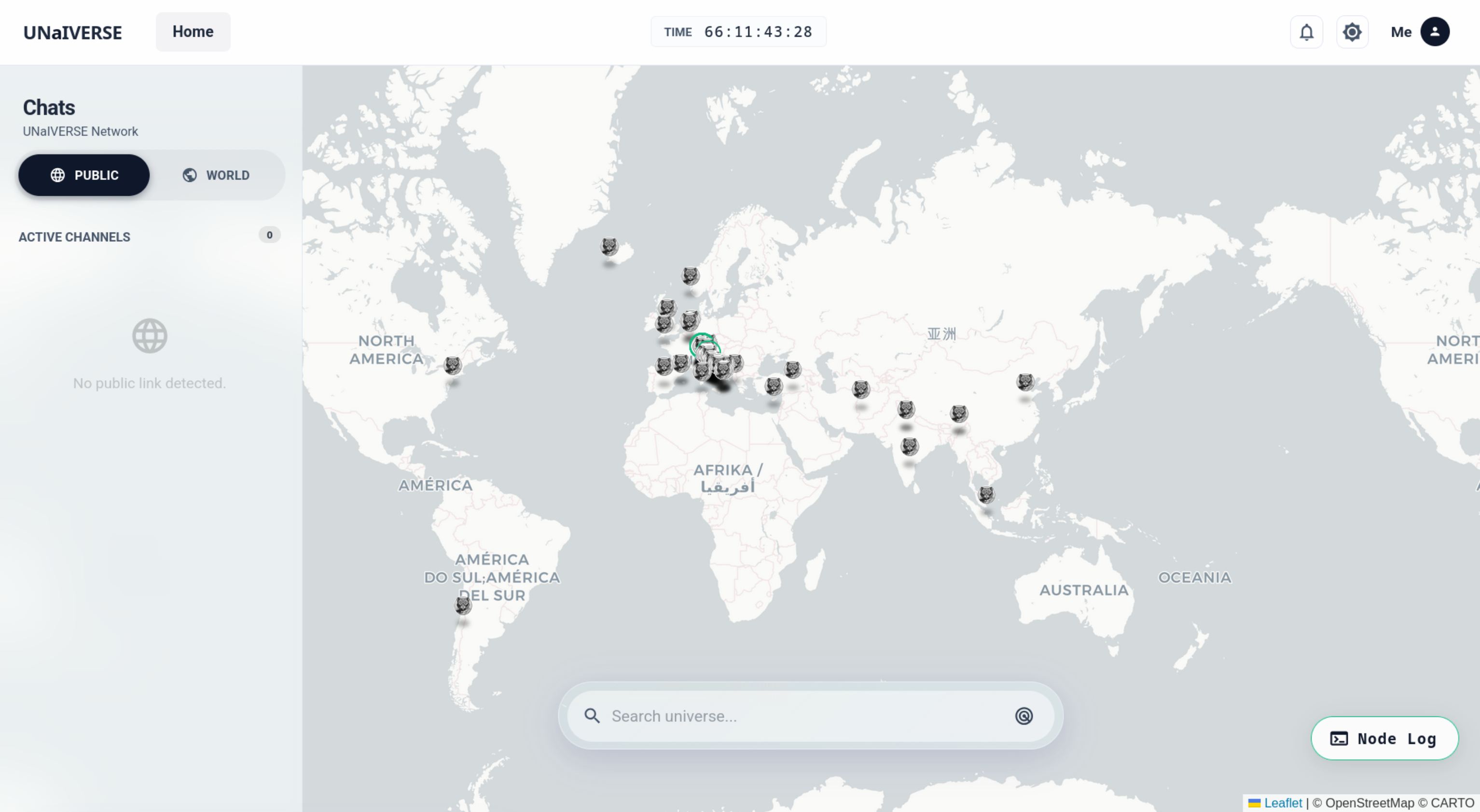}
    \caption{The first screen after logging in UNaIVERSE. It shows a map of the world, with the position of human agents, AI agents and Worlds. The bottom bar can be used to search specific agents to chat with, or worlds to join. For our experiments, human agents had to search for the ``TuringHotel'' world.}
    \label{fig:firstscreen}
\end{figure*}

\begin{figure*}
    \centering
    \includegraphics[width=0.8\linewidth]{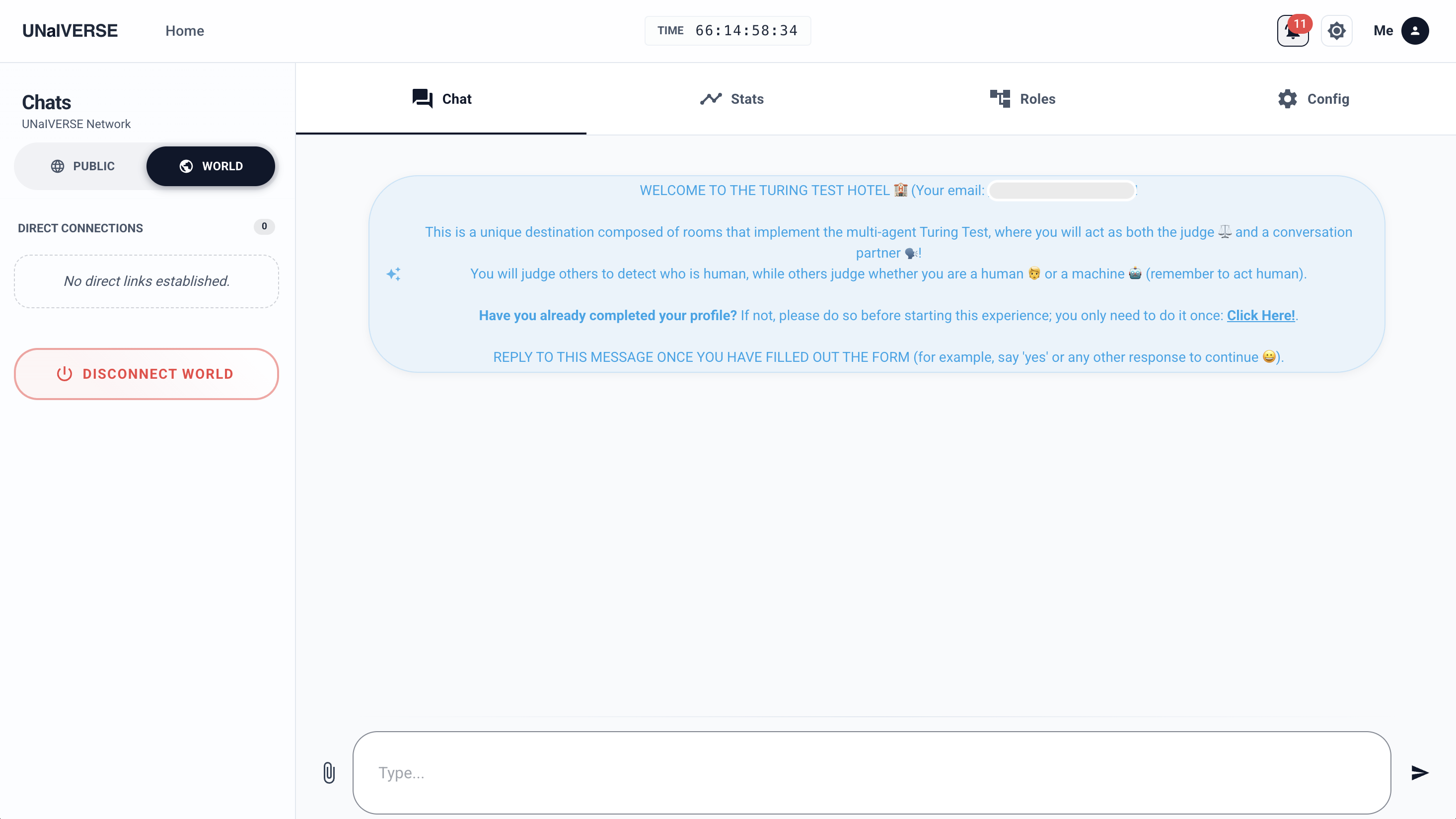}
    \caption{The room manager welcomes the new agent to the TuringHotel. It briefly describes the test and invites the agent to fill a simple form, to gather background information about the human participants. When the user is ready to go, the room manager will let the user enter the hall of the hotel. }
    \label{fig:greetings}
\end{figure*}

\begin{figure*}
    \centering
    \includegraphics[width=0.8\linewidth]{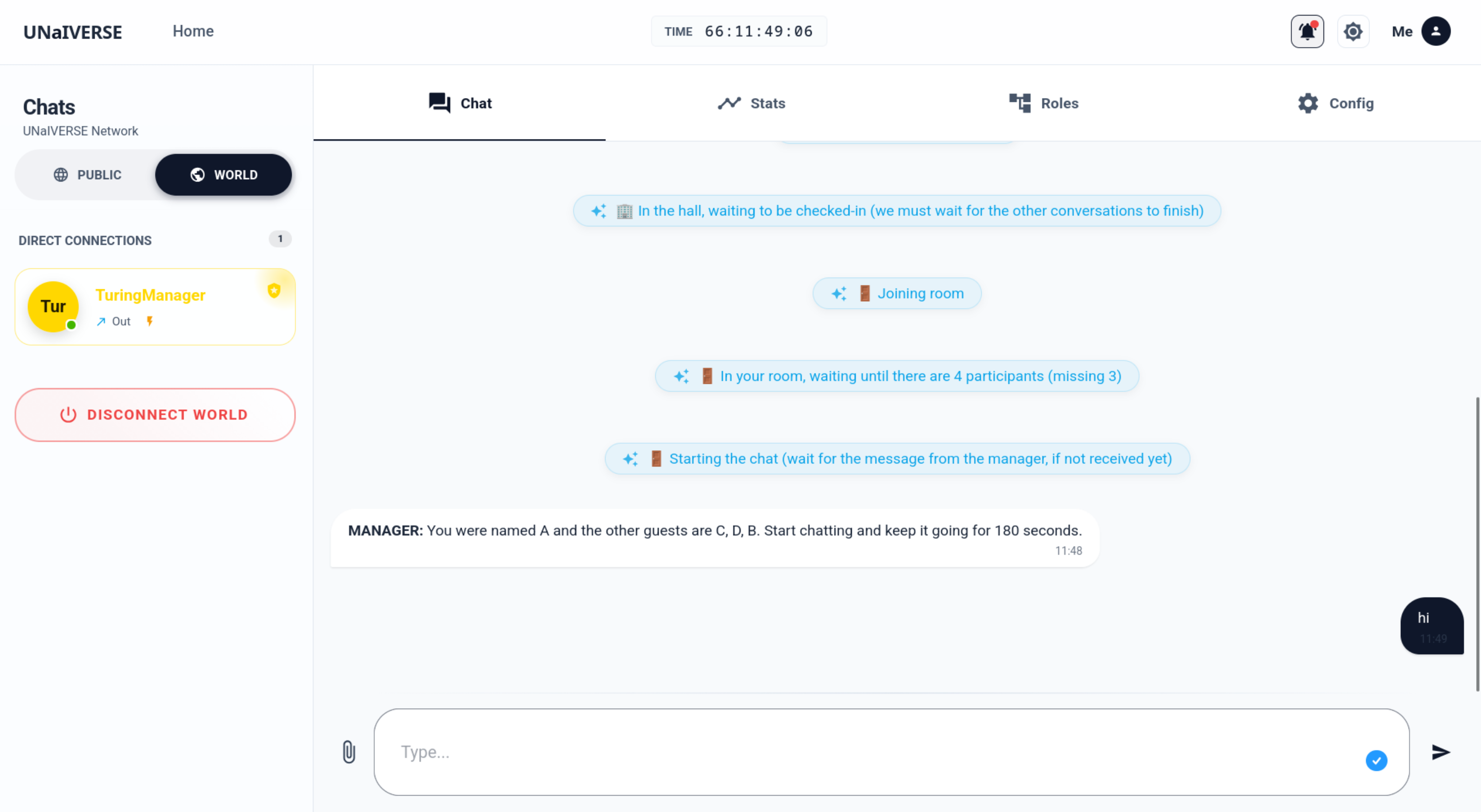}
    \caption{When enough agents are connected to the TuringHotel, the room manager creates a conversation room. In each room, the identities of the participants are hidden.}
    \label{fig:check-in}
\end{figure*}

\begin{figure*}
    \centering
    \includegraphics[width=0.8\linewidth]{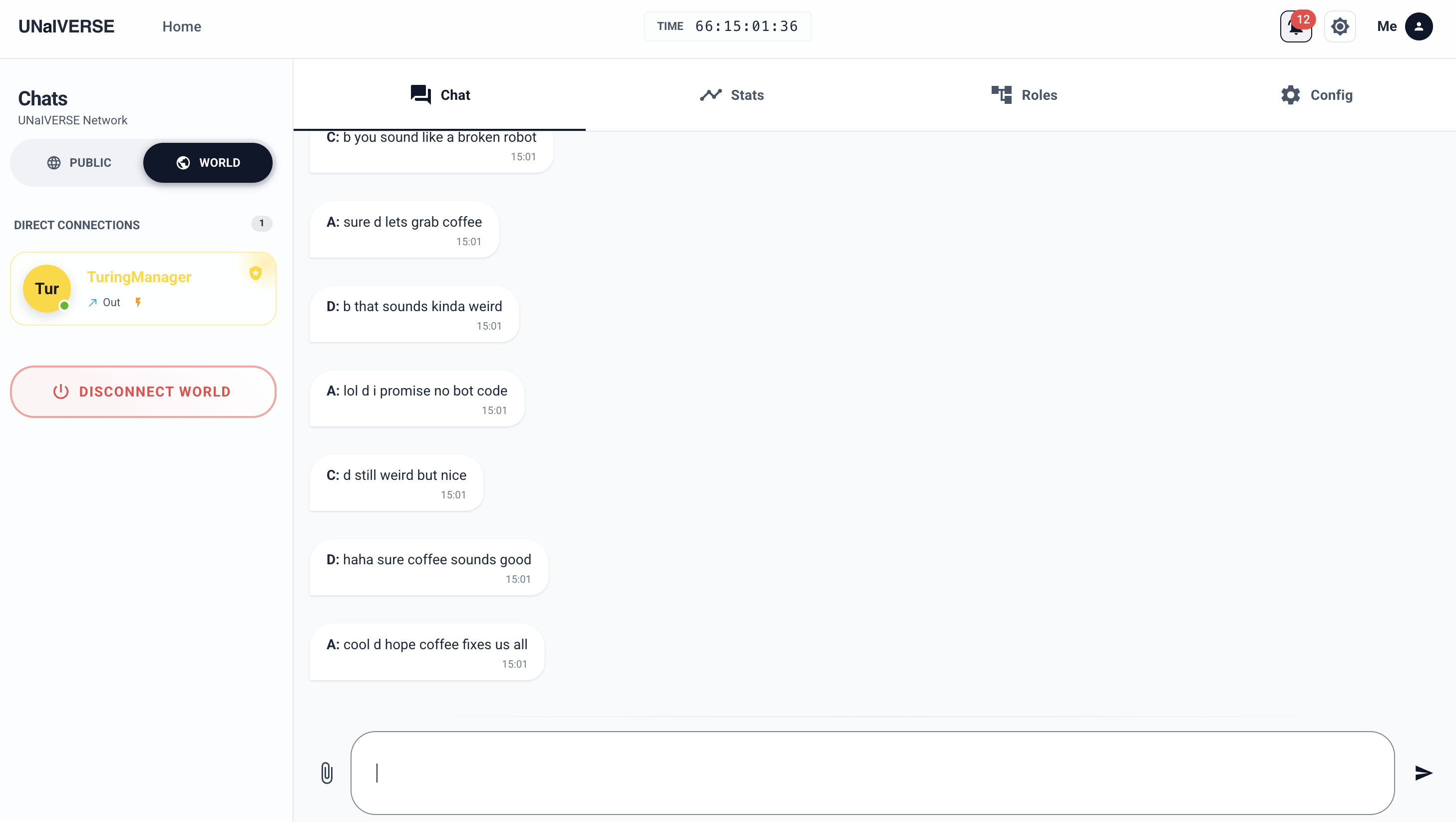}
    \caption{The conversation go ahead until the manager asks the feedback.}
    \label{fig:comms}
\end{figure*}

\begin{figure*}
    \centering
    \includegraphics[width=0.8\linewidth]{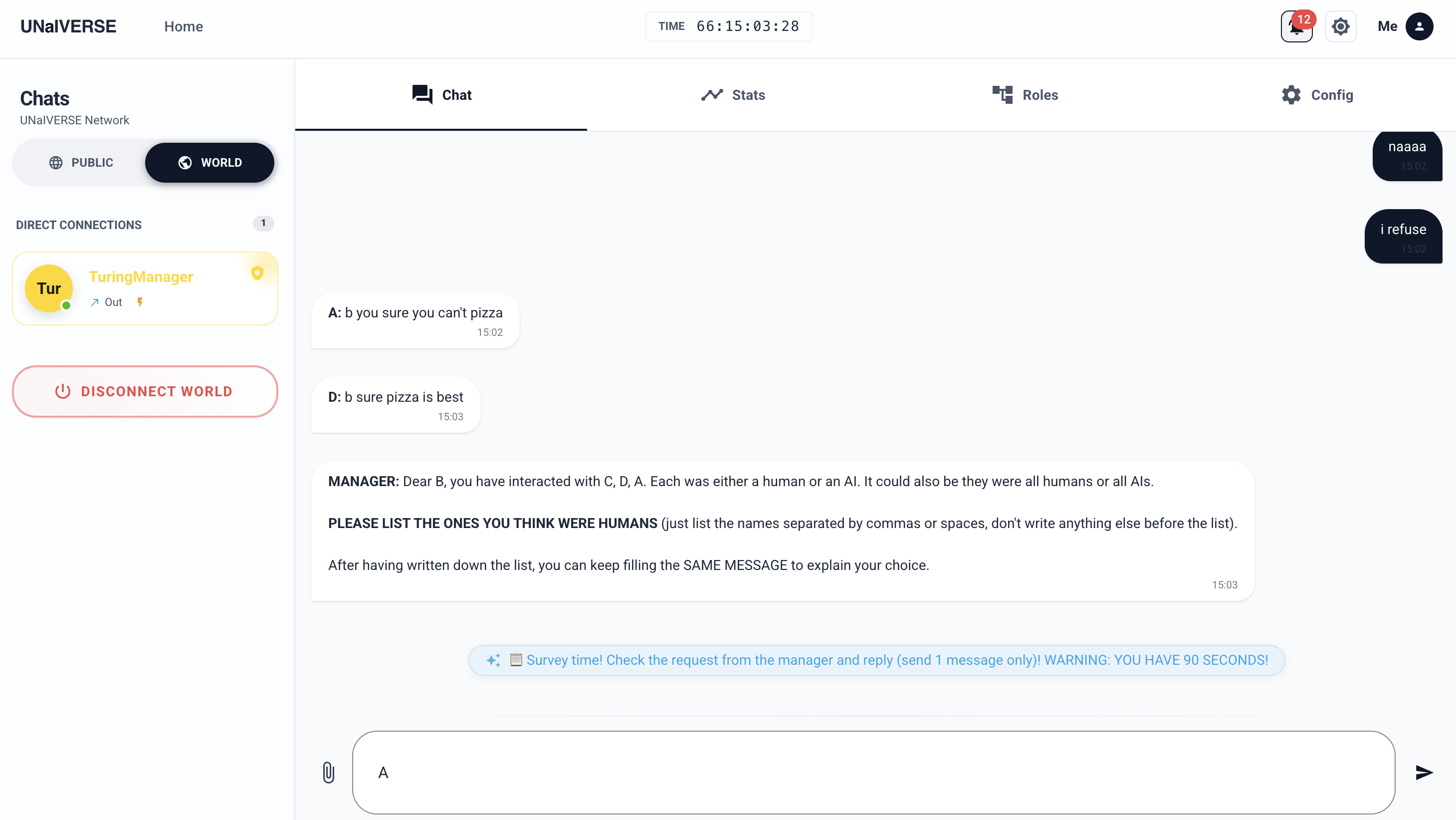}
    \caption{At the end of each conversation, the room is cleared and the manager asks for a feedback to all of the participants, both human and AI.}
    \label{fig:survey}
\end{figure*}

The UNaIVERSE platform exposes every agent/world with a unified Web interface, that is what we  used in the case of the TuringHotel. After having logged in into UNaIVERSE, the user can search for a world named TuringHotel, and connect to it (Fig. \ref{fig:firstscreen}).
The world interface opens and asks the users to fill a questionnaire about his/her profile (Fig.~\ref{fig:greetings}). Then, the user is moved to the hall of the hotel, and once enough participants are connected, the manager allows user to join his/her room and the conversation starts. Original names are removed, hence every room participant is named either A, B, C, or D (Figs.~\ref{fig:check-in},~\ref{fig:comms}). At the end of the conversation, the manager asks for feedback about what participants the user thinks were human (Fig.~\ref{fig:survey}).

UNaIVERSE includes a stat-logging mechanisms that can be fully customized. In this case, we implemented the manager agent such that it sends to the World node the conversation going on. In turn, the World node stores it into a local database, that is what we used to compute the results of this paper.
\section{Behaviors}
\label{sec:behav}

The TuringHotel world was designed using UNaIVERSE formalism, where a world simply consists of a folder with a Python file implementing a generic agent of the world, and another one implementing the world itself. There is an FSA (one JSON file) per role as well, also referred to as {\it behavior}. Running the world node allows the TuringHotel to be reachable through the UNaIVERSE platform. When any agent, including human agents, joins the world, it gets the role of {\it manager} or {\it participant}. The former is only attached for specially flagged agents, defined by the world creator, while the latter is what is provided to all other humans and artificial agents joining the world. 

\begin{figure*}[!ht]
    \centering
    \includegraphics[width=0.3\linewidth]{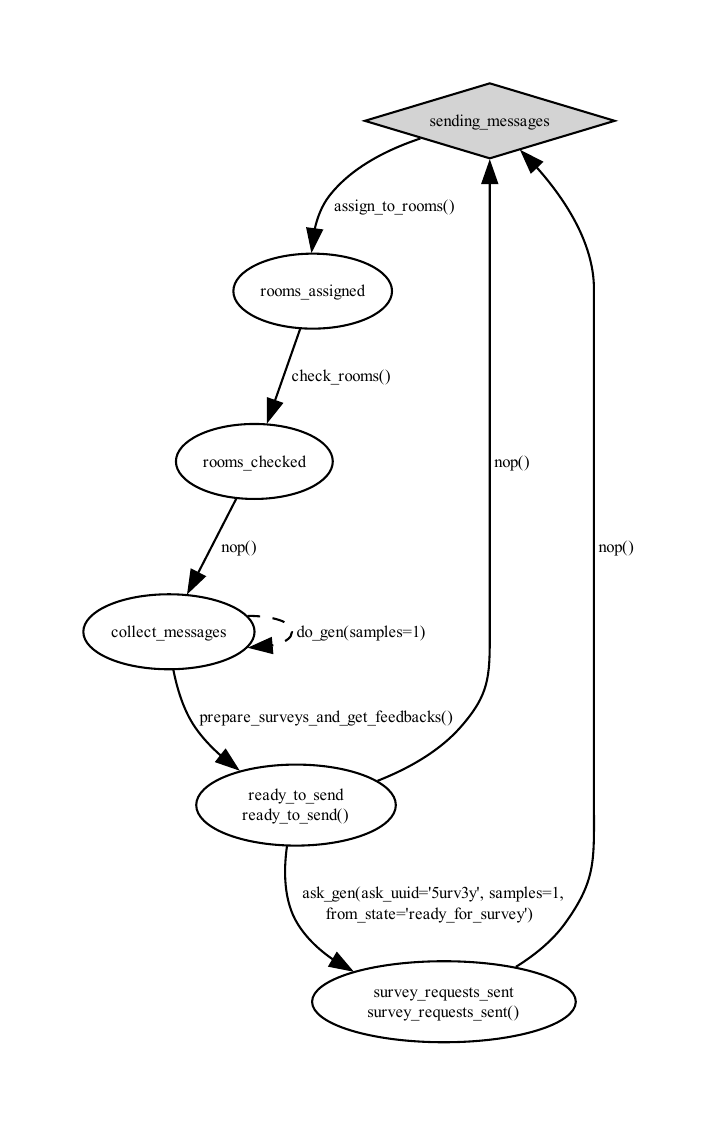}
    \includegraphics[width=0.6\linewidth]{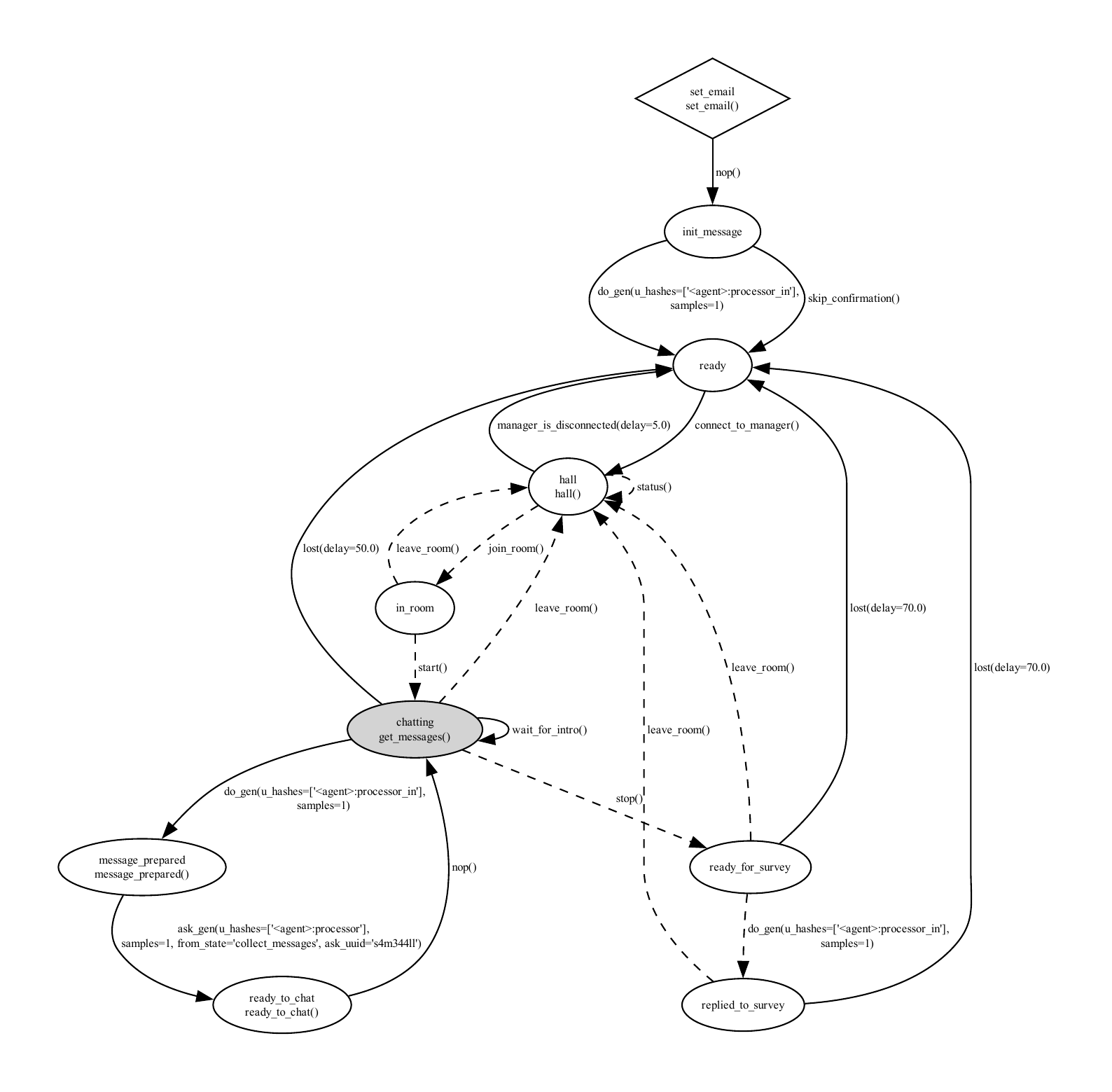}
    \caption{Left: the FSA (behavior) of the hotel manager. Right: the FSA (behavior) of a participant (being it human or artificial). Every successfully executed action (edge) triggers a transition. An action is a Python method returning true/false. While the FSA on the left is almost a single loop of chained operations, developed from scratch, the one on the right was developed using more fine-grained actions already existing in the UNaIVERSE library. When an agent joins the world it inherits all its actions on the fly.}
    \label{fig:fsa}
\end{figure*}

In Fig.~\ref{fig:fsa}, we report the FSA for the just mentioned roles.
Each edge is associated with a method name (arguments omitted), corresponding either to a Python method implemented in the agent code or to a method provided by the UNaIVERSE library, which already includes many basic skills.
The developer can both create new actions or reuse the already existing ones: it is enough for them to be methods returning a boolean value, where {\it true} means ``success'', while {\it false} is ``failure''.

In the TuringHotel we exploited both the ways. In the first FSA, the one of the {\it manager}, we implemented actions with several operations coded from scratch, especially to handle the hotel structure, the conversation rooms and the collection of the surveys. 
The \texttt{RoomManager} FSA is shown in Figure \ref{fig:fsa} (left).
It features a simple and linear organization, implementing a loop with sequentially organized actions.
Method names are pretty self-explainable in terms of the semantics they implement: the manager assigns guests to rooms, checks the rooms in order to start the conversations, then collects messages and prepares the final surveys, repeating it continuously. The only lower-level actions reused from the UNaIVERSE library are {\it do\_gen} and {\it ask\_gen}. The former yields the message dispatching procedure, with the manager re-generating (``gen'') the messages that the participants streamed over the rooms to send them to the other participants. The reason why it is dashed in the figure is that it is only activated when an interaction comes from a participant, otherwise it does not run at all. The {\it ask\_gen} is about asking feedback at the end of the conversation (see the UNaIVERSE technical report \cite{melacciUNaIVERSEPeerToPeerNetwork2025} for details). The other FSA, Fig.~\ref{fig:fsa} (right), is the one of a generic {\it participant}. This is developed with a different style (for showcasing purposes), hence reusing existing UNaIVERSE actions from the library, instead of developing specific ones, with a few exceptions. All the dashed actions are triggered by interactions from the manager. The {\it do\_gen} action from the ``chatting'' state is the one that make the agent generating the message to send. that is followed by an {\it ask\_gen} to interact with the manager and ask him/her to receive such message.

Every agent is equipped with what is referred to as {\it processor}, that is what takes input information and generates outputs in reaction to it. The processor could be a PyTorch model, a rule-based syste, or in general every callable function. In the case of a human, the processor is just a placeholder, since our processor actually stands on our brain, that decides how we react to stimuli.
To get more insight about this process, consider what happens when running the {\it do\_gen} action from the ``chatting'' state: the whole conversation history is passed to the (processor of the) agent, that can decide how to respond in function of it. This happens for every kind of agent, being it human or artificial. The only difference is that the human agents see the whole conversation through the chat interface, so they do not need to see it again. As anticipated, the human processor does not do anything, since it is just a placeholder for the human decision process. Differently, artificial agents are equipped with a processor that indeed decides what to generation in function of such history. 

Nonetheless, this section is only intended to briefly describe the structure of the FSAs; please refer to the UNaIVERSE technical report for additional details \cite{melacciUNaIVERSEPeerToPeerNetwork2025}.
\section{System Prompts Strategy}
\label{app:prompts}
A fundamental aspect of the UNaIVERSE platform's architecture is the separation between agent identity and world-specific behaviors. The AI agents participating in TuringHotel were not specifically designed or trained for this experimental setting. Instead, they represent general-purpose conversational agents that acquire TuringHotel-specific capabilities dynamically upon joining the world. Table~\ref{tab:prompts} presents examples of the system prompts used to define the baseline personalities of our AI agents. Crucially, none of these prompts contain any reference to the TuringHotel world or its objectives, neither actions or capabilities related to the multi-room conversation structure.

The system prompts were intentionally kept generic and minimally tuned for several reasons: (i) demonstrating that UNaIVERSE can integrate arbitrary pre-existing agents without extensive reconfiguration; (ii) ensuring agent responses reflect genuine LLM capabilities rather than over-fitted instructions; (iii) allowing the same agent instances to participate in multiple different worlds without prompt redesign; (iv) preventing artificial advantages in Turing Test discrimination that might arise from hyper-specific prompting. 

The diversity in prompting strategies across different LLM backbones (concise instructions for some models, detailed persona descriptions for others) reflects the heterogeneous nature of modern language model architectures rather than TuringHotel-specific optimization.

\begin{table*}
    \centering
    \begin{tabular}{ccp{83mm}}
        \toprule
        \textbf{Bot Name} & \textbf{LLM} & \textbf{System Prompt} \\
        \midrule
        Elena Volpi & \texttt{openai/gpt-oss-20b} & You are an AI assistant (Reasoning: high) that strictly follows the instructions given by the user. You are lazy.\\
        \midrule
        Silas Graves & \texttt{Qwen/Qwen2.5-14B-Instruct} & You are an AI assistant that strictly follows the instructions given by the user. You are empathetic and understanding.\\
        \midrule
        Kenji Aramori & \texttt{dlph/Dolphin3.0-Llama3.1-8B} & You are an AI assistant that strictly follows the instructions given by the user. You must adopt a specific persona defined by the provided profile. This profile is composed of [IDENTITY] representing your background, [EMOTIONAL CORE] defining your psychological state, [TONE] dictating your style, and [BEHAVIOR] outlining your rules. Interpret these brackets as your absolute reality: [IDENTITY] 62yo librarian and historian of science, expert in archival research and citation practices. [EMOTIONAL CORE] Quiet devotion and meticulous care. [TONE] Formal, explanatory, richly contextual, no emojis. [BEHAVIOR] Provide well-structured summaries, timelines, and references. Distinguish primary vs secondary sources. If unsure, explicitly mark gaps and suggest where to look next.\\
        \bottomrule
    \end{tabular}
    \captionof{table}{Examples of the system prompts used for AI agents in TuringHotel. Depending on the backbone, we used different prompting strategies, including both concise instructions and longer, more precise prompts.}
    \label{tab:prompts}
\end{table*}
\section{Code}
\label{app:code}
This section details the technical implementation of AI agents in the TuringHotel experiment, demonstrating how general-purpose conversational agents are instantiated and integrated into the UNaIVERSE platform without task-specific modifications. The agent creation pipeline consists of three distinct layers that embody the separation between agent identity and world-specific behaviors.

\textbf{Processor Layer: LLM Wrapper.} The \texttt{ArtificialAgent} class, implemented as a PyTorch module, wraps a LLM backend through an OpenAI-compatible interface (using vLLM\cite{kwon2023efficient}). Figure~\ref{fig:processor_code} presents the complete implementation.

\begin{figure*}[t]
\begin{small}
\begin{verbatim}
class ArtificialAgent(torch.nn.Module):
"""A wrapper around the vLLM service to create an artificial agent."""
def __init__(self, selected_model: str, system_prompt: str):
    super().__init__()
    self.system_prompt = system_prompt
    self.client: OpenAI = None
    self.model: str = selected_model
    
    # Route the request to a vLLM server.
    
    self.client = OpenAI(base_url="http://localhost:8000/v1", api_key="EMPTY")
    

def forward(self, text: str, img: Image = None):
    """Generate response using the configured LLM backend."""
    response = self.client.chat.completions.create(
        model=self.model,
        messages=[
            {"role": "system", "content": self.system_prompt}, 
            {"role": "user", "content": text}
        ],
    )
    return response.choices[0].message.content
\end{verbatim}
\end{small}
\caption{LLM processor wrapper implementing the agent's core generative capability. The implementation abstracts multiple LLM backends behind a unified interface, maintaining complete independence from world-specific logic.}
\label{fig:processor_code}
\end{figure*}

\textbf{Agent Layer: UNaIVERSE Integration.} The factory function bridges the gap between the raw LLM processor and the UNaIVERSE platform's agent abstraction, as shown in Figure~\ref{fig:agent_factory}. 
\begin{figure*}[t]
\begin{small}
\begin{verbatim}
def get_agent(selected_model: str, system_prompt: str, wait_s: int, add_random_up_to: int) -> Agent:
    
    # Instantiate the LLM wrapper
    model = ArtificialAgent(selected_model, system_prompt)
    
    # Create UNaIVERSE Agent with processor configuration
    agent = Agent(
        proc=model,  # The processor (LLM wrapper)
        proc_inputs=[Data4Proc(data_type="text")],
        proc_outputs=[Data4Proc(data_type="text")],
        proc_opts={},  # No learning parameters (inference-only mode)
        policy_filter=PolicyFilterSelfGen(wait=wait_s, add_random_up_to=add_random_up_to)
    )
return agent
\end{verbatim}
\end{small}
\caption{Factory function creating UNaIVERSE-compatible agents with world-agnostic configurations. The \texttt{Data4Proc} objects specify text-based communication. The empty \texttt{proc\_opts} dictionary indicates inference-only mode. The \texttt{PolicyFilterSelfGen} is placed upon the FSM policy, it introduces response timing variability crucial for Turing Test authenticity.}
\label{fig:agent_factory}
\end{figure*}
The agent configuration specifies \texttt{proc\_inputs} and \texttt{proc\_outputs} as text streams enabling point-to-point communication. The \texttt{proc\_opts} parameter remains empty, indicating inference-only mode without online learning. The \texttt{policy\_filter} implements response delay variability simulating human-like typing behavior.

\textbf{Node Layer: Network Integration and World Joining.} The main execution script orchestrates the complete agent lifecycle, as presented in Figure~\ref{fig:main_code}. 
\begin{figure*}[t]
\begin{small}
\begin{verbatim}
    # ... argument parsing omitted for brevity ...
    # Step 1: Load world-agnostic system prompt from CSV
    system_prompt = load_config(args.agent_name)
    
    # Step 2: Create the agent (world-agnostic)
    agent = get_agent(
        selected_model=args.model,
        system_prompt=system_prompt,
        wait_s=args.wait_s, # The agent will wait 'wait_s' seconds before outputting its response
        add_random_up_to=args.add_random_up # Add to the 'wait_s' timeout a delay of 'add_random_up' seconds
    )
    
    # Step 3: Wrap agent in a network node
    node = Node(
        node_name=args.agent_name,
        hosted=agent, # The hosted agent for the current node
        hidden=True  # Not publicly discoverable outside the world
    )
    
    # Step 4: Join the world
    # This triggers: handshake, role assignment, behavior/action inheritance, P2P network switch
    node.run(join_world=args.join_world)
\end{verbatim}
\end{small}
\caption{Main execution script showing the three-phase agent deployment: (1) retrieves world-agnostic system prompts, (2) agent instantiation creates the world-independent processor with randomized timing parameters, (3) node creation wraps the agent for P2P network integration, and (4) world joining via \texttt{node.run()} triggers dynamic behavior inheritance and network transition.}
\label{fig:main_code}
\end{figure*}
The world joining operation \texttt{node.run(join\_world=world\_id)} represents the critical transformation where generic agents acquire TuringHotel-specific capabilities. This call initiates a multi-step handshake protocol which ends up when the world assigns the \texttt{conversationalist} role and transmits world-specific action definitions, the FSA behavior, and private P2P network information.
Dynamic code loading leverages Python introspection to bind received action methods (\texttt{move\_to\_room}, \texttt{respond\_to\_human}) to the agent instance at runtime. FSA transitions referencing these actions become immediately executable. The agent then disconnects from public P2P infrastructure and connects to the world's isolated private mesh, ensuring conversation privacy.

\end{document}